\documentclass[lettersize,journal]{IEEEtran}
\usepackage{amsmath,amsfonts}
\usepackage{algorithmic}
\usepackage{array}
\usepackage[caption=false,font=normalsize,labelfont=sf,textfont=sf]{subfig}
\usepackage{textcomp}
\usepackage{stfloats}
\usepackage{url}
\usepackage{verbatim}
\usepackage{graphicx}
\usepackage{algorithm}
\usepackage{algorithmic}
\usepackage{multirow}
\makeatletter

\newcommand{\Rmnum}[1]{\expandafter\@slowromancap\romannumeral #1@}
\makeatother
\def\BibTeX{{\rm B\kern-.05em{\sc i\kern-.025em b}\kern-.08em
    T\kern-.1667em\lower.7ex\hbox{E}\kern-.125emX}}
\usepackage{balance}
\begin{document}
\title{DEPFusion: Dual-Domain Enhancement and Priority-Guided Mamba Fusion for UAV Multispectral Object Detection}
\author{Shucong Li, Zhenyu Liu, \textit{Member, IEEE}, Zijie Hong, Zhiheng Zhou, \textit{Member, IEEE}, Xianghai Cao \textit{Member, IEEE}
\thanks{
This work has been submitted to the IEEE for possible publication.
Copyright may be transferred without notice, after which this version may
no longer be accessible.

Shucong Li, Zhenyu Liu and Zijie Hong are with the School of Information Engineering, Guangdong University of Technology, Guangzhou 510006, China (e-mail: 
    2112403242@mail2.gdut.edu.cn;
	zhenyuliu@gdut.edu.cn; 
	2112403242@mail2.gdut.edu.cn)

Zhiheng Zhou is with the 
School of Electronic and Information Engineering, South China University of Technology, Guangzhou 510630, China (e-mail: zhouzh@scut.edu.cn)

Xianghai Cao is with the School of Artificial Intelligence, XiDian University. Xi'an  710071, China (e-mail: caoxh@xidian.edu.cn)

The code will be available.
}}

\markboth{Journal of \LaTeX\ Class Files,~Vol.~18, No.~9, September~2020}%
{How to Use the IEEEtran \LaTeX \ Templates}

\maketitle

\begin{abstract}
Multispectral object detection is an important application for unmanned aerial vehicles (UAVs). However, it faces several challenges. First, low-light RGB images weaken the multispectral fusion due to details loss. Second, the interference information is introduced to local target modeling during multispectral fusion. Third, computational cost poses deployment challenge on UAV platforms, such as transformer-based methods with quadratic complexity. 
To address these issues, a framework named DEPFusion consisting of two designed modules, Dual-Domain Enhancement (DDE)  and Priority-Guided Mamba Fusion (PGMF) , is proposed for UAV multispectral object detection. 
Firstly, considering the adoption of low-frequency component for global brightness enhancement and frequency spectra features for texture-details recovery,
DDE module is designed with Cross-Scale Wavelet Mamba (CSWM) block and Fourier Details Recovery (FDR) block. 
Secondly, considering guiding the scanning of Mamba from high priority score tokens, which contain local target feature, 
a novel Priority-Guided Serialization is proposed with theoretical proof. Based on it, PGMF module is designed for multispectral feature fusion, which enhance local modeling and reduce interference information.
Experiments on DroneVehicle and VEDAI datasets demonstrate that DEPFusion achieves good performance with state-of-the-art methods. 
\end{abstract}


\section{Introduction}
\IEEEPARstart{U}{namaned} aerial vehicles (UAVs) play an important role in civil and public fields, including pest controlling, transportation monitoring and relief work etc. \cite{bridging} In addition, 
object detection is one of the most basic application of UAV, which depends on the remote sensing images provided by RGB camera and infrared imaging sensor. The RGB images contain color and texture-details 
about the real world, but their performance is limited by the low-light condition or overexposure. Instead, the infrared images provide stable capture of thermal data and target's outlines under any visual conditions, complementing the advantages offered by RGB images.

Therefore, many studies have focused on designing UAV-based object detection frameworks. According to the chosen sensor modality, these methods could be categorized into RGB-based and multispectral object detection.

The RGB-based object detection method relies on the color and texture-details provided by cameras on the UAVs. Sun et.al
\cite{sun} modify YOLOv8 \cite{yolov8} which exists an issue of losing small object, obtaining higher accuracy of detection. Xiao et.al \cite{xiao} propose a lightweight framework for small object detection based on multi-scale feature fusion. Lin et al. \cite{Lin} propose GE-FSOD to address the challenge of limited labeled data and poor generalization in remote sensing object detection. Liang et al. \cite{Liang} propose the FS-SSD model with spatial context analysis to tackle small object detection in UAV imagery. Yu et al. \cite{Yu} propose ARPN to address the challenges of detecting tiny persons in UAV imagery, where targets are small in complex backgrounds. Overall, these works advance to RGB-based object detection for UAVs application.

The multispectral object detection introduces infrared images as auxiliary branch to overcome the low-light issue of RGB images. TarDAL \cite{tardal} introduces generative adversarial network to reserve the common feature of two modalities in the fused images. SuperYOLO \cite{superyolo} adopts input-level fusion of RGB and infrared images, enhancing feature representation by the super-resolution branch, while feature-wise fusion operates on multi-stage backbone features. To align fusion with downstream tasks, DetFusion \cite{detfusion} guides training through detection loss derived from fused images.
Sun et.al \cite{dronevehicle} design UA-CMDet and a novel RGB-Infrared drone dataset named DroneVehicle. However, UA-CMDet adopts simple feature concatenation to achieve multispectral fusion. Zhang et.al \cite{tracking} introduce complementary image and discriminative feature fusion to make full use of the effective information between the two modalities and filter out redundant features. To address inaccurate cross-modal fusion,  $
\text{C}^2\text{Former}
$ \cite{c2former} employs CNNs for feature extraction, while utilizes a Transformer-based Inter-modality Cross-Attention module to generate aligned and complementary features. Inspired by the linear complexity of Mamba, DMM \cite{dmm} designs disparity-guided mamba fusion to adaptively merge features and achieve lightweight object detection.

\begin{figure}[t]
	\centering
	\includegraphics[width=3.3in]{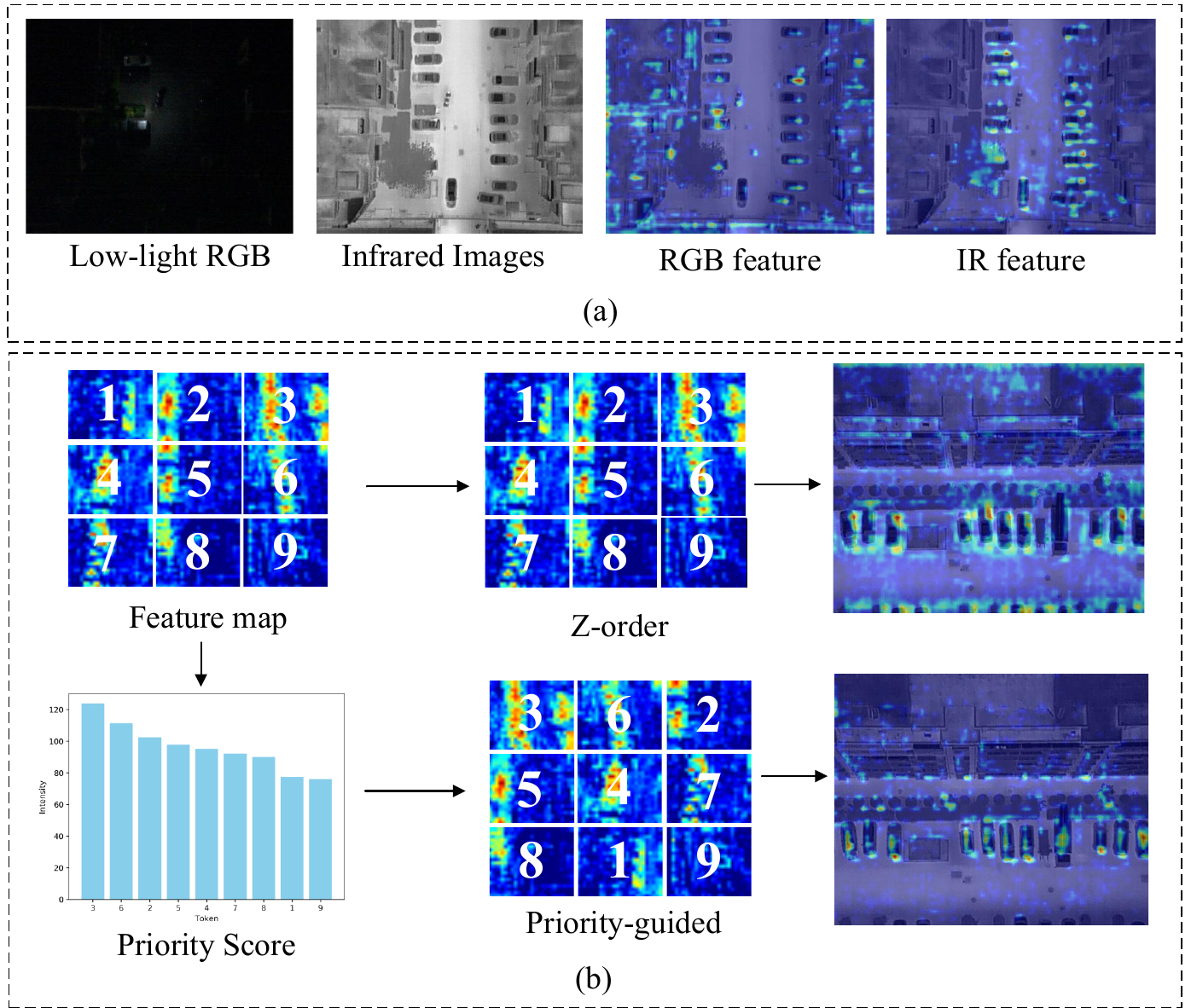}
	\caption{The modality feature difference and the proposed priority-guided scanning strategy. (a) The feature difference between low-light RGB images and infrared images. (b) The comparison between Z-order scanning and our Priority-guided scanning. \label{ques}
	}
\end{figure}
However, the above methods also face three  challenges: 1) Although the infrared images provide supplementary information about scenarios, the low-light RGB images weaken multispectral fusion due to details loss, causing the obvious differences in their feature maps. As shown in Fig \ref{ques} (a).  2) In the fusion stage, the local small target modeling
are interfered with redundant information. Specially, as shown in Fig \ref{ques} (b), the Mamba-based fusion methods like DMM adopts Z-order scanning focuses on global modeling and suppresses the local dependencies of 2D images. Therefore, the feature map contains noise. 3) The transformer-based methods like $
\text{C}^2\text{Former}$ has 
large computational costs for detection task.

To address these challenges, DEPFusion is proposed with Dual-domain Enhancement and Priority-guided Mamba Fusion for multispectral object detection. Firstly, to enhance low-light RGB images, DDE Module is designed, which contains CSWM and FDR block. Specially, CSWM block contains Cross-Scale Mamba Scanning for the low-frequency components from 2D discrete wavelet transform (2D-DWT) to enhance the global brightness of images. FDR block constructs Spectrum Recovery Network (SRN) to enhance the frequency spectra features to recover the texture-details. Secondly, Priority-Guided Serialization is proposed with theoretical proof, which adopting priority scores to guide the scanning of Mamba to prioritize tokens containing local target features. Based on Priority-Guided Serialization, 
PGMF module is designed to enhance the local target modeling for multispectral fusion. PGMF constructs Priority Score Network (PSN) to obtain the priorities of tokens from feature differences.  
Experiment results on DroneVehicle and VEDAI dataset demonstrate that DEPFusion achieves lightweight computation and performs well on object detection.

Overall, our contributions could be summarized as follow:
\begin{itemize}
	\item[$\bullet$] DEPFusion is proposed, a novel multispectral object detection framework with low-light RGB images enhancement and priority-guided Mamba multi-modality fusion.
	
	\item[$\bullet$] DDE is designed and contains CSWM and FDR block. CSWM block improves the global brightness and FDR block recovers the texture-details of low-light RGB images. 
	
	\item[$\bullet$] To the best of our knowledge, it is the first time to propose Priority-Guided Serialization with theoretical proof. Based on it, PGMF module is designed to enhance local target modeling and reduce the interference information for multispectral feature fusion.

	\item[$\bullet$] The experiment results on DroneVehicle dataset and VEDAI dataset show that DEPFusion  performs well on object detection and achieves lightweight computation.

\end{itemize}

\section{Related Work}
\subsubsection{Multispectral Object Detection}
The multispectral usually includes visible and infrared images, both of which could provide complementary information under unstable lighting conditions for object detection. Existing multispectral fusion mechanism could be classified into pixel-wise fusion and feature-wise fusion.

Pixel-wise fusion aggregates two spectral data into a whole images ahead of schedule, and feeds the images to the detector. Cao et.al \cite{multi} propose a local–global adaptive dynamic learning method to fuse multi-modal data. To ensure the fusion network is task-optimized, DetFusion \cite{detfusion} employs the detection loss computed from the fused images as guidance during training. Employing input-level fusion of RGB and IR images. Feature-wise fusion focuses on the features from different stages of backbone network. These stage features are fused and fed to the detection head. TSFADet \cite{tsr} designs an alignment module to predict modality deviations with subsequent feature map calibration. Zhao et.al \cite{removal} propose a coarse-to-fine framework to remove redundant information and select features for fusion dynamically.
To resolve modality calibration and fusion inaccuracies, $
\text{C}^2\text{Former}
$ \cite{c2former} introduced an inter-modal cross-attention module, while its Adaptive Feature Sampling module mitigated computational costs from global attention.

However, these works focuses on the scenarios under normal lighting conditions, ignoring the application on the low-light scenarios. Moreover, they lack the attention to reduce the impact of redundant information for local small target detection.

\subsubsection{Low-light Images Enhancement}
There are many frameworks having been  designed for enhancing the low-light images, which lacks enough texture-details due to under exposed characteristic. Brightness enhancement in FourLLIE \cite{fourll} is achieved by dual-stage estimation of amplitude transformation maps within the frequency domain. DiffLL \cite{diffll} implements sequential denoising refinements in diffusion models to generate realistic details for low-light image enhancement task, highlighting their practical viability.
\cite{lowlightaaai} proposes a framework with guidance of generative perceptual priors obtained from vision-language models, it shows good generalization on real-world data. WaveMamba \cite{wavemamba} combines wavelet transformation and Mamba mechanism to achieve low-light images restoration. These works enhance the texture-details of instances in low-light images, which gives important help to the detection task.

\begin{figure*}[t]
	\centering
	\includegraphics[width=7in]{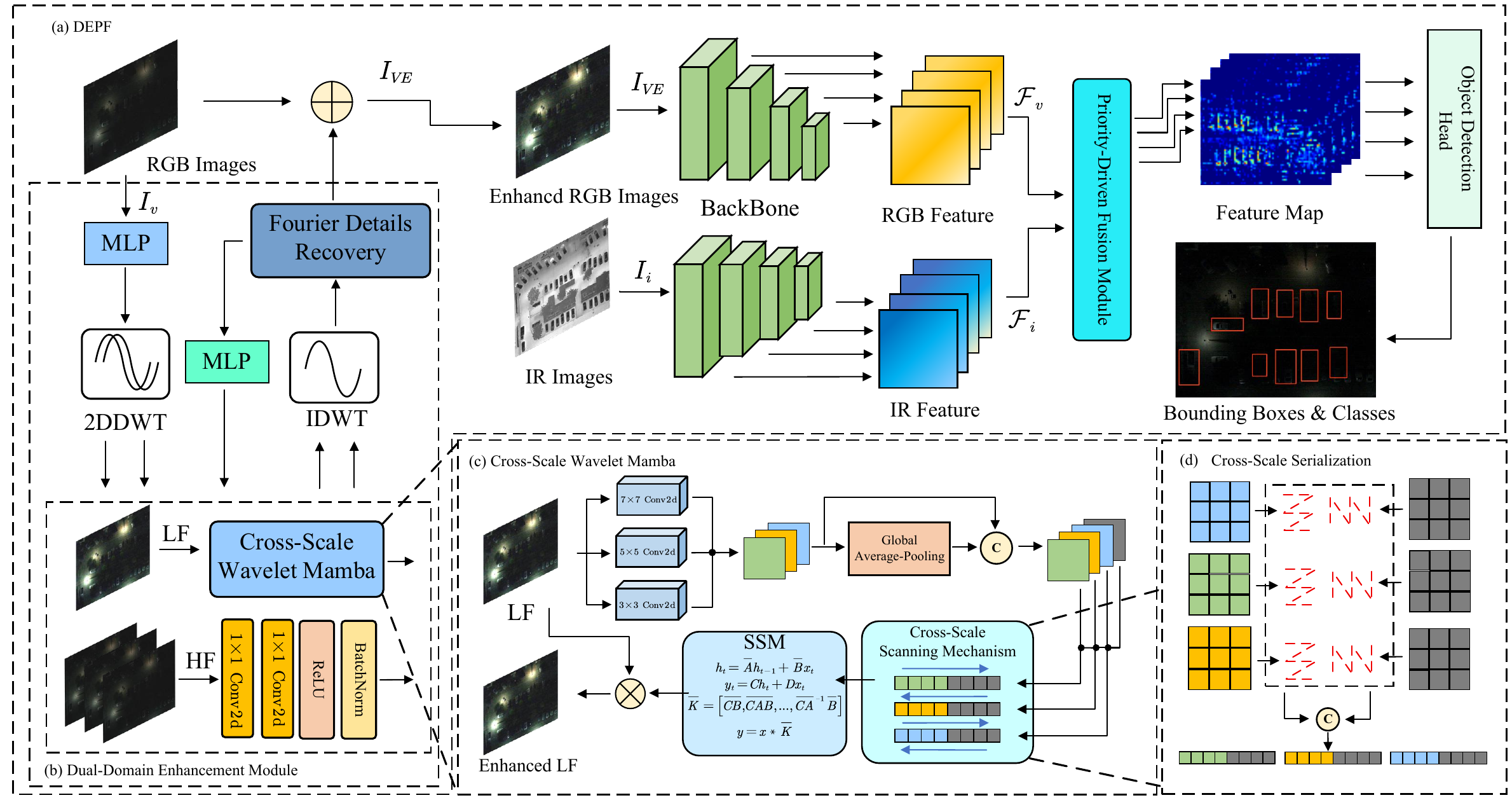}
	\caption{The overall architecture of DEPFusion. (a) is the pipeline of DEPFusion, (b) is the architecture of DDE module, while (c) is the processing of CSWM block including the Cross-Scale Scanning Mechanism. (d) shows the Cross-Scale Serialization in the CSWM block. \label{pipeline}
	}
\end{figure*}

\subsubsection{Vision Mamba}
Mamba \cite{mamba} mechanism has been paid attention to computer vision field because it can model long-distance dependencies and maintain linear computational complexity. Many vision mamba works has developed and performed well in the vision researches. The visual state-space model VMamba \cite{vmamba}  preserves SSM’s global receptive field capability but operates at linear computational cost, enabling order-of-magnitude speed gains. Vim \cite{vim} proposes a new vision backbone with bidirectional scanning mamba, This architecture embeds positional cues into sequences while compacting representations through bidirectional state-space compression. Vision mamba is widely adopted for images classification tasks, like SpectralMamba \cite{spectralmamba}, $\text{S}^2\text{Mamba}$ \cite{s2mamba} etc., and images segmentation tasks like $\text{RS}^3\text{Mamba}$ \cite{rs3mamba},  
VM-Unet \cite{vmunet} etc. These works could achieve good performance with low computation complexity.

However, most of vision mamba works adopts Z-order serialization for scanning the images, which focuses on global context and suppresses the local dependencies.
Therefore, for the remote sensing images with many small objects, the interference information brings degradation to local targets modeling in the multi-modality fusion stage under the Z-order serialization.

\section{Method}
\subsection{Dual-Domain Mamba Enhancement Module}
The low-frequency component in wavelet-domain contains most of information about low-light RGB images, while the high-frequency component only contains small parts of texture-details \cite{wavemamba}. Therefore, the low-frequency component could be adopted to enhance the overall brightness
of images. 

For the texture-details, it is worth noting that frequency domain has good global modeling capacity. The amplitude spectrum contains the energy of details while the phase spectrum contains the structure of the instances in the images. Therefore, these two spectra could be adopted to recover the texture details of images.

Following the above ideas, DDE module is designed for enhancing the low-light RGB images, which is shown in Fig \ref{pipeline}. DDE module consists of two parts, the first part is CSWM block, which is designed for low-frequency component enhancement. The second part is FDR block, which is designed to enhance the texture details of images.

\subsubsection{Cross-Scale Wavelet Mamba Block}
For the low-light RGB remote sensing images $I_v\in \mathbb{R}^{B\times C\times H\times W}$, they are decomposed by the N-level Haar wavelet, yielding low-frequency components $LL$ and high-frequency components $\left[ HL,LH,HH \right] _n$:
\begin{equation}\label{eqn-1}
	LL,\left[ HL,LH,HH \right] _n,...,\left[ HL,LH,HH \right] _1=DWT\left( I_v \right) 
\end{equation}
where $DWT()$ denotes 2D-DWT operator.

The $LL$ is enhanced by CSWM block, which is designed by adopting the strong context-aware properties of Mamba \cite{mamba}. To ensure that the  local objects dependencies are enhanced, $LL$ is conveyed to three convolution layers with kernel sizes $i\in  \left[ 3,5,7 \right]$ to obtain multi-scale feature and global feature:
\begin{equation}\label{eqn-2}
	f_i=DWConv\left( LL,i \right) \forall i\in \left\{ 3,5,7 \right\} 
\end{equation}
\begin{equation}\label{eqn-3}
	f_g=GAP\left( f_3,f_5,f_7 \right) 
\end{equation}
where $DWConv()$ denotes depthwise convolution, $GAP()$ denotes global average pooling operator.

For learning the context information, the Cross-Scale Scanning Mechanism is designed.
Concretely, every scale feature $f_i$ and the global feature $f_g$ are flatten and concatenated together so that the relative scale information could be learned. To adequately exploit the latent features, the forward scanning and backward scanning are set up simultaneously. Therefore, the cross-scale scanning sequences could be described as:
\begin{equation}\label{eqn-4}
	f_{cross}^{i}=flatten\left( f_i \right) +flatten\left( f_g \right) 
\end{equation}
\begin{equation}\label{eqn-5}
	f_{seq}^{i}=f_{cross}^{i}+\overline{f_{cross}^{i}}\ \forall i\in \left\{ 3,5,7 \right\}
\end{equation}
where $flatten()$ denotes flatting operator and $\overline{f_{cross}^{i}}$ denotes the reverse sequence of $f_{cross}^{i}$.
After the SSM module $SSM()$ , the enhanced $LL$ is obtained by:
\begin{equation}\label{eqn-6}
	LL_e=LL\times \sum_i^{}{SSM\left( f_{seq}^{i} \right)}\ \forall i\in \left\{ 3,5,7 \right\} 
\end{equation}

The high-frequency components are passed through MLP and join in wavelet reconstruction together with $LL_e$:
\begin{equation}\label{eqn-7}
	I_{ve}^{n}=IDWT\left( LL_e,\left[ HL,LH,HH \right] _n \right) \ \forall n\in \left[ 0,N \right] \cap \mathbb{Z}
\end{equation}
\begin{figure}[t]
	\centering
	\includegraphics[width=3.3in]{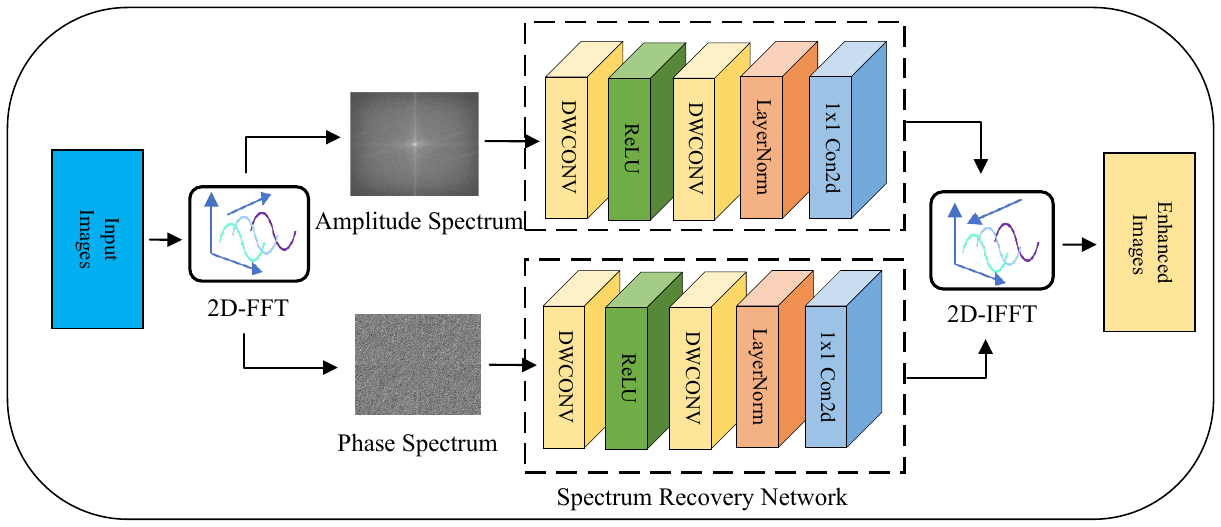}
	\caption{The processing of FDR block. It enhances the amplitude and phase features by spectrum recovery network. \label{DDE}
	}
\end{figure}

\subsubsection{Fourier Details Recovery}
The enhanced images $I_{ve}^{n}$ also lacks texture-details after low-frequency enhancement. Therefore, $I_{ve}^{n}$ is conveyed to FDR module for recovering most of texture-details. The structure of FDR module could be seen in Fig \ref{DDE}. FDR contains spectrum recovery networks which could enhance the details of amplitude spectrum (AS) and phase spectrum (PS) obtained by 2D-FFT. The enhanced MS and PS are then transformed into time-domain to obtain rich texture-details images $I_{VE}$. This process could be described as:
\begin{equation}\label{eqn-8}
	\begin{split}
		AS&=abs \left( FFT\left( I_{ve}^{n} \right) \right) \\
		PS\ &=\ \arctan \left( \frac{imag\left( FFT\left( I_{ve}^{n} \right) \right)}{real\left( FFT\left( I_{ve}^{n} \right) \right)}, \right)\\
		AS_e&=SRN\left( AS \right) \\
		PS_e&=SRN\left( PS \right) \\
		I_{VE}&=IFFT\left( AS_e,PS_e \right) \\
	\end{split}
\end{equation}	
where $SRN()$ denotes the spectrum recovery network, which consists of two DWCONV layers, one ReLU layer, one LayerNorm layer and one Linear layer.  $FFT()$ denotes fast 2D-FFT operator, and $IFFT()$ denotes the inverse transformation of 2D-FFT. $I_{VE}$ should be conveyed to CSWM block if it is not in the last level of 2D-DWT.

\begin{figure*}[t]
	\centering
	\includegraphics[width=7in]{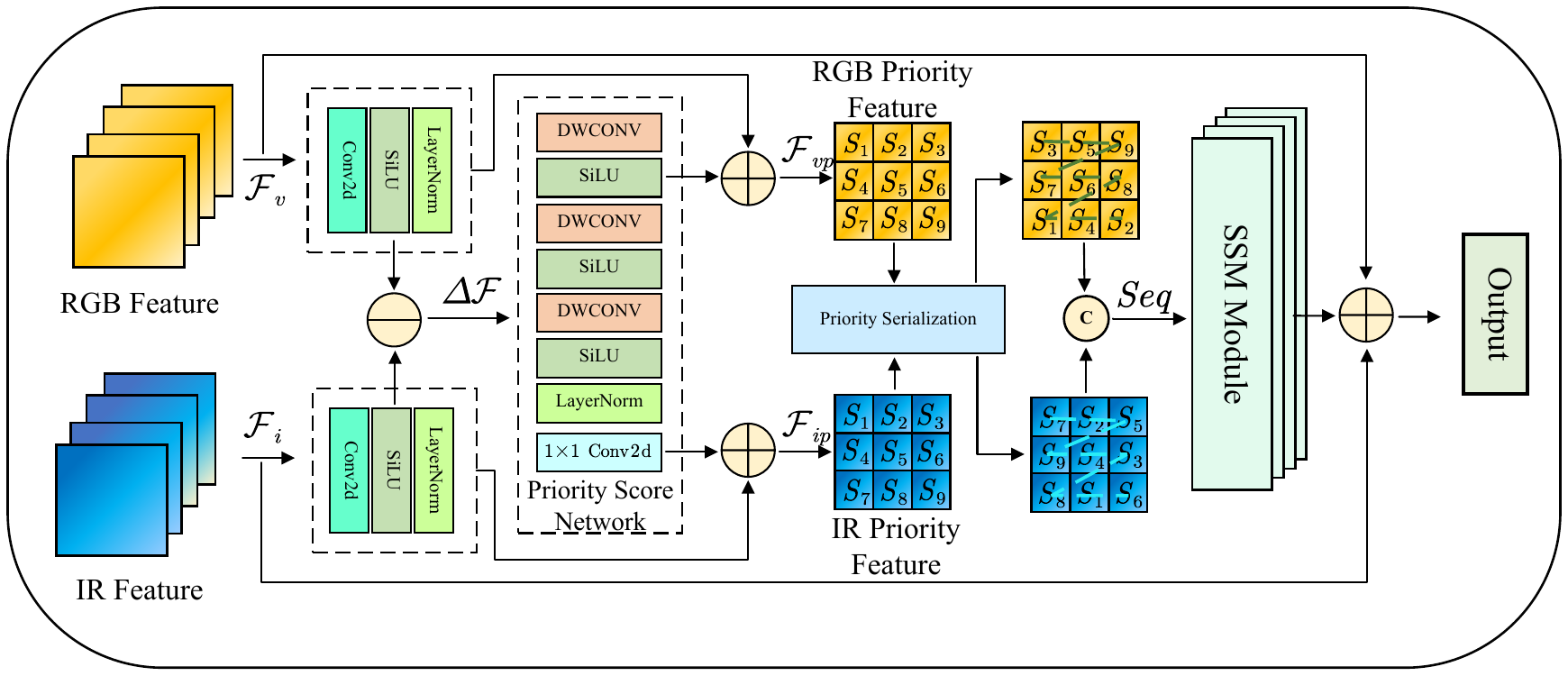}
	\caption{The over architecture of PGMF module. The modalities features are transformed into sequences following the priority scores of tokens, which is obtained from Priority Score Network. \label{pm}
	}
\end{figure*}

\begin{algorithm}[t]
	\caption{Algorithm of the PGMF Module. 	\label{alg}}
	\textbf{Input}: RGB Feature: $\mathcal{F}_v$, Infrared Feature: $\mathcal{F}_i$ \\
	\textbf{Output}: Fusion Feature: $\mathcal{F}_{fus}$
	\begin{algorithmic}[1] 
		\STATE $\mathcal{F}_{vl}:\left( B,C,H,W \right) \gets Ref\left( \mathcal{F}_v \right) $
		\STATE $
		\mathcal{F}_{il}:\left( B,C,H,W \right) \gets Ref\left( \mathcal{F}_i \right) 
		$
		\STATE $\varDelta \mathcal{F}:\left( B,C,H,W \right) \gets \mathcal{F}_{vl}-\mathcal{F}_{il}$
		\STATE $PMat:\left( B,C,H,W \right) \gets PSN\left( \varDelta \mathcal{F} \right)$
		\STATE$\mathcal{F}_{vp}:\left( B,C,H,W \right) \gets \mathcal{F}_{vl}+PMat$
		\STATE $\mathcal{F}_{ip}:\left( B,C,H,W \right) \gets \mathcal{F}_{il}+PMat$
		\STATE $\mathcal{F}_{vp}:\left( B,C,H \times W \right) \gets Split\left( \mathcal{F}_{vp} \right) 
		$
		\STATE$
		\mathcal{F}_{ip}:\left( B,C,H \times W \right) \gets Split\left( \mathcal{F}_{ip} \right)$ 
		\STATE $index_v\gets argSort\left( \mathcal{F}_{vp} \right) 
		$
		\STATE$
		index_i\gets argSort\left( \mathcal{F}_{ip} \right) $
		\STATE$\text{Initialize}\ \text{modality}\ \text{sequences}\ Seq_v, Seq_i $
		\FOR{ $t_v$, $t_i$  in $
			index_v$,$index_i$}
		\STATE $token_v\gets \mathcal{F}_{vl}\left[:,:, t_v \right] 
		$
		\STATE$
		token_i\gets \mathcal{F}_{il}\left[:,:,t_i \right] 
		$
		\STATE$
		Seq_v\gets Concat\left( Seq_v,token_v \right) 
		$
		\STATE$
		Seq_i\gets Concat\left( Seq_i,token_i \right)
		$
		\ENDFOR
		\STATE $Seq:\left( B,C,2 \times H\times W \right) \gets Concat\left( Seq_v,\overline{Seq_i} \right) 
		$
		\STATE$
		\mathcal{F}_{pv}:\left( B,C, H\times W \right), \mathcal{F}_{pi}:\left( B,C, H\times W \right),
		\gets SSM\left( Seq \right) 
		$
		\STATE$
		\mathcal{F}_p:\left( B,C,H,W \right) \gets Dropout\left( \mathcal{F}_{pv} + \mathcal{F}_{pi}  \right) 
		$
		\STATE$
		\mathcal{F}_{fus}:\left( B,C,H,W \right) \gets \mathcal{F}_v+\mathcal{F}_i+Reshape\left(\mathcal{F}_p \right)$
	\end{algorithmic}
\end{algorithm}
\subsection{Priority-Guided Mamba Fusion Module}
For the feature maps of remote sensing images, there is an obvious difference in the proportion of local target features and background features. However, the Z-order serialization focuses on global modeling, which brings redundant information to weaken the local targets modeling. Therefore, if the scanning prioritizes the feature of local targets, the impact of redundant information is lessened, and the original global feature is retained. The fused feature map with obvious targets details could be obtained by scanning the priority sequences of two modalities. In the later section, we conduct theoretical proof about the advantage of Priority-guided for Mamba modeling. 

Following the above ideas, Priority-Guided Mamba Fusion  (PGMF) module is designed to achieve high effective RGB-IR feature fusion, which is shown in Fig \ref{pm}.  

\subsubsection{The Process of PGMF Module}
The input feature $\mathcal{F}_v$ and $\mathcal{F}_i$ are conveyed to small MLPs to reflect into latent space. Moreover, it is worth noting that not only does the feature difference $\varDelta \mathcal{F}$ contain location information of the targets, but also show the regions that be focused mistakenly.
Therefore, $\varDelta \mathcal{F}$ is  conveyed to Priority Score Network to obtain priority matrices $PMat$. Both of two modalities could obtain their priority features by adding with $PMat$:
\begin{equation}\label{eqn-9}
	\begin{split}
		\varDelta \mathcal{F}&=\mathcal{F}_{vl}-\mathcal{F}_{il}\\
		PMat&=PSN\left( \varDelta \mathcal{F} \right) \\
		\mathcal{F}_{vp}&=\ \mathcal{F}_{vl}+PMat\\
		\mathcal{F}_{ip}&=\,\,\mathcal{F}_{il}+PMat\\
	\end{split}
\end{equation}
where $\mathcal{F}_{vl}$ and $\mathcal{F}_{il}$ denote the latent representations of RGB feature map and IR feature map. $PSN()$ denotes the Priority Score Network, which consists of three DWCONV-SiLU layers, one LayerNorm layer and one Linear layer.

The serialization sets up to obtain the sequences by following the descent priority scores, which separates the target from interference information. Therefore, the scanning starts from the targets and ends up with the background information. 

For the reason that the parameter renewing of $A$ matrix of Mamba mechanism relies on the historical information, and it is necessary to improve the contributions of local object feature. The fusion of two sequences is conducted by the format of $Seq$ in Eq (10):
\begin{equation}\label{eqn-10}
	\begin{split}
		Seq_v&=Ser\left( F_{vl},argSort\left( F_{vp} \right) \right) \\
		Seq_i&=Ser\left( F_{il},argSort\left( F_{ip} \right) \right) \\
		Seq\ &=\ Seq_v+\overline{Seq_i}
	\end{split}
\end{equation}
Where $Ser()$ denotes serialization operator and $argSort()$ denotes obtaining the sequences after sorting.

Two sequences are fused together and conveyed to SSM module, which contains forward scanning and backward scanning. The input features of two modalities are added with priority feature by residual connection:
\begin{equation}\label{eqn-11}
	\begin{split}
		\mathcal{F}_p&=SSM\left( Seq \right) \\
		\mathcal{F}_{fus}&=\mathcal{F}_p+\mathcal{F}_v+\mathcal{F}_i
	\end{split}
\end{equation}

The pseudocode PGMF module could be seen in \textbf{Algorithm 1}.

\subsubsection{The Theoretical Proof of Priority-Guided Serialization}
As known to all, the current output of Mamba \cite{mamba} depends on the historical information. In addition, the selective learning mechanism is achieved by decaying the hidden state dynamically, which allows Mamba to capture long-range dependencies.  Therefore, by proofing that the decaying is related to the position of token, 
we could proof the advantage of Priority-Guided Serialization, and the fusion way expanded from it.  We define the following symbols firstly:
\begin{equation*}
	\begin{split}
		&\overline{A},\overline{B},\overline{C}:\ \text{The\ parameter\ matrices\ of\ SSM\ in\ discrete\ states }
		\\
		&A:\ \text{Parameter\ }\overline{\text{A}}\ \text{in\ the\ continuous\ state}\\
		&h:\ \text{The\ hidden\ state}
		\\
		&y:\ \text{The\ output\ of\ SSM}
		\\
		&x:\ \text{The\ input\ of\ SSM} \\
		&\varDelta :\ \text{Discretization\ step\ size}
	\end{split}
\end{equation*}
The hidden state and the output of Mamba could be describe as:
\begin{equation}\label{eqn-12}
	\begin{split}
		h_t&=\overline{A}h_{t-1}+\overline{B}x_t
		\\
		y_t&=\overline{C}h_t
	\end{split}
\end{equation}
where $\overline{A}\in \mathbb{R}^{n\times n},\overline{B}\in \mathbb{R}^{n\times 1},\overline{C}\in \mathbb{R}^{1\times n},x\in \mathbb{R}^{1\times n}$. Therefore, if we define a sequence 
\begin{equation}\label{eqn-13}
	x=\left[ x_1,x_2,...,x_t,...,x_T \right] 
\end{equation} 
the $t-th$ output could be described as:
\begin{equation}\label{eqn-14}
	h_t=\sum_{i=0}^t{\overline{A}^i\overline{B}x_{t-i}}
\end{equation}
for the contribution of $i-th$ token to the $t-th$ output, it could be obtained from Eq (14):
\begin{equation}\label{eqn-15}
	\lVert h_{t}^{i} \rVert =\lVert \left( \prod_{k=i+1}^t{A_k} \right) B_ix_i \rVert
\end{equation}
According to Holder Inequation, the contribution has supremum:
\begin{equation}\label{eqn-16}
	\lVert h_{t}^{i} \rVert =\lVert \left( \prod_{k=i+1}^t{\overline{A}_k} \right) \overline{B}_ix_i \rVert \le \lVert \prod_{k=i+1}^t{\overline{A}_k} \rVert \lVert \overline{B}_ix_i \rVert
\end{equation}
According to ZOH principle, we could calculate the eigenvalue of $\overline{A}$:
\begin{equation}\label{eqn-17}
	\lambda \left( \overline{A}_k \right) =e^{\varDelta \lambda \left( A \right) _k}<1
\end{equation}
since $\overline{A}=-e^{\ln A}=-A<0$.
The spectral radius of $\overline{A}$ satisfies:
\begin{equation}\label{eqn-18}
	\rho_k =\underset{k}{\max}\left[ \lambda \left( \overline{A}_k \right) \right] <1
\end{equation}
Following Geland Theorem and Eq (18), we could know that:
\begin{equation}\label{eqn-19}
	\underset{\left| i-t \right|\rightarrow \infty}{\lim}\lVert \prod_{k=i+1}^t{\overline{A}_k} \rVert =0 
\end{equation}
The limiting of the item in Eq (16) could be obtained:
\begin{equation}\label{eqn-20}
	\underset{\left| i-t \right|\rightarrow \infty}{\lim}\lVert \prod_{k=i+1}^t{\overline{A}_k} \rVert \lVert B_ix_i \rVert =0
\end{equation}
For the $i-th$ token, if it is far from the $t-th$ hidden state, the contribution has infimum:
\begin{equation}\label{eqn-21}
	\underset{\left| i-t \right|}{\text{inf}}\left( \lVert h_{t}^{i} \rVert \right) =0
\end{equation}
Following Eq (20), we could obtain the range of the contribution value:
\begin{equation}\label{eqn-22}
	0\le \underset{\left( t-i \right) \rightarrow \infty}{\lim}\lVert h_{t}^{i} \rVert \le \underset{\left( t-i \right) \rightarrow \infty}{\lim}\lVert \prod_{k=i+1}^t{A_k} \rVert \lVert B_ix_i \rVert =0
\end{equation}
Following Squeeze Theorem, we could conclude that:
\begin{equation}\label{eqn-23}
	\underset{\left( t-i \right) \rightarrow \infty}{\lim}\lVert h_{t}^{i} \rVert =0
\end{equation}
From Eq (23) we conclude that $A$ matrix decays the information of early tokens and capture new information of late tokens. However, the parameter renewing of $A$ matrix is impacted by the historical information such as the initial tokens.
Therefore, for the multispectral fusion sequence, the tokens locating on the head and tail are important. This requirement guides us to construct the fusion sequence by:
\begin{equation}\label{eqn-24}
	X=x_{vp}+\overline{x_{ip}}
\end{equation}
where $x_{vp}$ and $x_{ip}$ denote the priority sequences of two modalities. Eq (24) shows that the local target feature are stored in the head and tail of sequence.
Under the forward and backward scanning of SSM, the features of two modalities are fused complementarily. 

\subsubsection{The Computational Complexity of PGMF Module}
The computational complexity of PSN could be described as:
\begin{equation}
	O_{psn}=O\left( \sum_{}^{}{\left( H\times W\times K^2\times C_{in}\times C_{out} \right)} \right)
\end{equation}
where $H$,$W$ denote the height and width of images respectively, $K$ denotes the size of convolution kernel, $C_{in}$, $C_{out}$ denote the number input channels and output channels respectively. Notably,  $K$, $C_{in}$ and $C_{out}$ are constant parameters. Therefore, Eq (25) could be simplified as:
\begin{equation}
	O_{psn}=O\left( N \right)
\end{equation}
where $N$ denotes the number of pixels and $N=H\times W$.

The sorting contributes the computational complexity on the Priority-Guided serialization. Notably, the Radix Sort is adopted by PyTorch, which has linear complexity. For the SSM module, it also has linear complexity:
\begin{equation}
	\begin{split} 
		O_{pri}&=O\left( N \right)\\
		O_{ssm}&=O\left( N \right)
	\end{split} 
\end{equation}

Therefore, our PGMF module improves the performance of detector and maintains linear complexity, comparing with conventional SSM fusion with Z-order serialization.
\subsubsection{The Back-Propagation of PGMF Module}
In this section, we will derive the gradient of PGMF module during back-propagation. We define the following symbols firstly:
\begin{equation*}
	\begin{split}
		&\mathcal{K}:\ \text{The\ kernel\ function\ of\ SSM}\\
		&x:\ \text{The\ input\ sequence\ of\ SSM}\\
		&Y:\ \text{The\ output\ of\ PGMF\ module}\\
		&\omega :\ \text{The\ parameters\ of\ }\mathcal{K}\\
		&f_v:\text{The\ feature\ extraction\ branch\ of\ RGB\ images} \\
		&f_i:\,\,\text{The\,\,feature\,\,extraction\,\,branch\,\,of\ IR\ images}\\
		&f_{psn}:\ \text{The\ Priority\ Score\ Network}\\
		&T:\ \text{The\ number\ of\ tokens}\\
		&\mathcal{L}:\ \text{The\ loss\ function}\\
		&\theta _v:\ \text{The\,\,parameter\,\,vector\,\,of\ }f_v
		\\
		&\theta _i:\,\,\text{The\,\,parameter\,\,vector\,\,of\,\,}f_i
		\\
		&\theta _{psn}:\,\,\text{The\,\,parameter\,\,vector\,\,of\,\,}f_{psn}
	\end{split}
\end{equation*}
The output of Mamba \cite{mamba} $y$ could be described by adopting convolution form:
\begin{equation}
	y=x\ast \mathcal{K}
\end{equation}
The sequence fed to SSM module is combined with the priority-guided sub-sequences of two modality. They are obtained from PSN, whose parameters are shared between two branches. Following Eq (28), This process could be described as:
\begin{equation}
	Y=\sum_{t=0}^T{\left\{ \mathcal{K}\left( \omega _t \right) \left[ f_v+f_{spn}\left( f_v \right) +f_i+f_{spn}\left( f_i \right) \right]_t \right\}}
\end{equation}
Therefore, the gradient vectors of $f_v$,$f_i$ and $f_{psn}$ could be obtained by the Chain Rule:
\begin{equation}
	\begin{split}
		&\nabla \mathcal{L}\left( \theta _v \right) =\sum_{t=0}^T{\left\{ \mathcal{K}\left( w_t \right) \left[ \nabla f_v\left( \theta _v \right) +\nabla f_{psn}\left( f_v \right) \nabla f_v\left( \theta _v \right) \right] _t \right\}}\\
		&\nabla \mathcal{L}\left( \theta _i \right) =\sum_{t=0}^T{\left\{ \mathcal{K}\left( w_t \right) \left[ \nabla f_i\left( \theta _i \right) +\nabla f_{psn}\left( f_i \right) \nabla f_i\left( \theta _i \right) \right] _t \right\}}\\
		&\nabla \mathcal{L}\left( \theta _{psn} \right) =2\sum_{t=0}^T{\left\{ \mathcal{K}\left( w_t \right) \left[ \nabla f_{psn}\left( \theta _{psn} \right) \right] _t \right\}}
	\end{split} 
\end{equation}
\subsection{Loss Function}
The loss function of DEPFusion  follows conventional object detection loss and classification loss.
Therefore, the Focal Loss \cite{celoss} function is adopted for the optimizing of object detection task:
\begin{equation}
	\mathcal{L}_{\det}=-\alpha _t\left[ p\left( 1-t \right) +t\left( 1-p \right) \right] ^{\gamma _t}\log \left( p_t \right)
\end{equation}
where $\alpha _t$ denotes the weight factor of class $t$, $p_t$ denotes the predicted value which represents the probability of class $t$.

The Smooth L1 loss \cite{l1loss} function is adopted  for the optimizing of regression task:
\begin{equation}
	\mathcal{L}_{reg}=\left\{ \begin{array}{l}
		\frac{\left| y_{gt}-y_{pred} \right|^2}{2},\ \left| y_{gt}-y_{pred} \right|<1\\
		\left| y_{gt}-y_{pred} \right|-\frac{1}{2},\ \left| y_{gt}-y_{pred} \right|\ge 1\\
	\end{array} \right.   
\end{equation}
where $y_{gt}$ and $y_{pred}$ denote the groundtruth value and predicted value respectively.

Therefore, the loss function of DEPFusion  could be described as:
\begin{equation}
	\mathcal{L}=\alpha \mathcal{L}_{det}+\beta \mathcal{L}_{reg}
\end{equation}
where $\alpha$ and $\beta$ denote the weights of loss item. During training phase, they are set to $\left\{ 1.0,1.0 \right\} $.

\begin{table*}[t]
	\centering
	\small
	\caption{Comparative experiment results on the DroneVehicle dataset. '-' represents that the result is not reported. \label{dv}}
	\renewcommand{\arraystretch}{0.6}
	\setlength{\tabcolsep}{2mm}{
	\begin{tabular}{cc|cc|cccccccccc|cccc}
		\hline
		\multicolumn{2}{c|}{\multirow{2}[1]{*}{Methods}} & \multicolumn{2}{c|}{\multirow{2}[1]{*}{Modality}} & \multicolumn{2}{c}{\multirow{2}[1]{*}{Car}} & \multicolumn{2}{c}{\multirow{2}[1]{*}{Truck}} & \multicolumn{2}{c}{\multirow{2}[1]{*}{Freight Car}} & \multicolumn{2}{c}{\multirow{2}[1]{*}{Bus}} & \multicolumn{2}{c|}{\multirow{2}[1]{*}{Van}} & \multicolumn{2}{c}{\multirow{2}[1]{*}{mAP@0.5}} & \multicolumn{2}{c}{\multirow{2}[1]{*}{mAP@0.5:0.95}} \\
		\multicolumn{2}{c|}{} & \multicolumn{2}{c|}{} & \multicolumn{2}{c}{} & \multicolumn{2}{c}{} & \multicolumn{2}{c}{} & \multicolumn{2}{c}{} & \multicolumn{2}{c|}{} & \multicolumn{2}{c}{} & \multicolumn{2}{c}{} \\
		\hline
		\multicolumn{2}{c|}{\multirow{2}[1]{*}{RetinaNet \cite{retinanet}}} & \multicolumn{2}{c|}{\multirow{2}[1]{*}{RGB}} & \multicolumn{2}{c}{\multirow{2}[1]{*}{78.5 }} & \multicolumn{2}{c}{\multirow{2}[1]{*}{34.4 }} & \multicolumn{2}{c}{\multirow{2}[1]{*}{24.1 }} & \multicolumn{2}{c}{\multirow{2}[1]{*}{69.8 }} & \multicolumn{2}{c|}{\multirow{2}[1]{*}{28.8 }} & \multicolumn{2}{c}{\multirow{2}[1]{*}{25.0 }} & \multicolumn{2}{c}{\multirow{2}[1]{*}{23.4 }} \\
		\multicolumn{2}{c|}{} & \multicolumn{2}{c|}{} & \multicolumn{2}{c}{} & \multicolumn{2}{c}{} & \multicolumn{2}{c}{} & \multicolumn{2}{c}{} & \multicolumn{2}{c|}{} & \multicolumn{2}{c}{} & \multicolumn{2}{c}{} \\
		\multicolumn{2}{c|}{\multirow{2}[0]{*}{$\text{R}^3\text{Det}$ \cite{r3det}}} & \multicolumn{2}{c|}{\multirow{2}[0]{*}{RGB}} & \multicolumn{2}{c}{\multirow{2}[0]{*}{80.3 }} & \multicolumn{2}{c}{\multirow{2}[0]{*}{56.1 }} & \multicolumn{2}{c}{\multirow{2}[0]{*}{42.7 }} & \multicolumn{2}{c}{\multirow{2}[0]{*}{80.2 }} & \multicolumn{2}{c|}{\multirow{2}[0]{*}{44.4 }} & \multicolumn{2}{c}{\multirow{2}[0]{*}{27.4 }} & \multicolumn{2}{c}{\multirow{2}[0]{*}{27.4 }} \\
		\multicolumn{2}{c|}{} & \multicolumn{2}{c|}{} & \multicolumn{2}{c}{} & \multicolumn{2}{c}{} & \multicolumn{2}{c}{} & \multicolumn{2}{c}{} & \multicolumn{2}{c|}{} & \multicolumn{2}{c}{} & \multicolumn{2}{c}{} \\
		\multicolumn{2}{c|}{\multirow{2}[0]{*}{KFIoU \cite{kfiou}}} & \multicolumn{2}{c|}{\multirow{2}[0]{*}{RGB}} & \multicolumn{2}{c}{\multirow{2}[0]{*}{76.3 }} & \multicolumn{2}{c}{\multirow{2}[0]{*}{50.0 }} & \multicolumn{2}{c}{\multirow{2}[0]{*}{33.2 }} & \multicolumn{2}{c}{\multirow{2}[0]{*}{78.2 }} & \multicolumn{2}{c|}{\multirow{2}[0]{*}{35.7 }} & \multicolumn{2}{c}{\multirow{2}[0]{*}{54.7 }} & \multicolumn{2}{c}{\multirow{2}[0]{*}{28.0 }} \\
		\multicolumn{2}{c|}{} & \multicolumn{2}{c|}{} & \multicolumn{2}{c}{} & \multicolumn{2}{c}{} & \multicolumn{2}{c}{} & \multicolumn{2}{c}{} & \multicolumn{2}{c|}{} & \multicolumn{2}{c}{} & \multicolumn{2}{c}{} \\
		\multicolumn{2}{c|}{\multirow{2}[0]{*}{RoITransformer \cite{roitrans}}} & \multicolumn{2}{c|}{\multirow{2}[0]{*}{RGB}} & \multicolumn{2}{c}{\multirow{2}[0]{*}{61.6 }} & \multicolumn{2}{c}{\multirow{2}[0]{*}{55.1 }} & \multicolumn{2}{c}{\multirow{2}[0]{*}{42.3 }} & \multicolumn{2}{c}{\multirow{2}[0]{*}{85.5 }} & \multicolumn{2}{c|}{\multirow{2}[0]{*}{44.8 }} & \multicolumn{2}{c}{\multirow{2}[0]{*}{61.6 }} & \multicolumn{2}{c}{\multirow{2}[0]{*}{32.9 }} \\
		\multicolumn{2}{c|}{} & \multicolumn{2}{c|}{} & \multicolumn{2}{c}{} & \multicolumn{2}{c}{} & \multicolumn{2}{c}{} & \multicolumn{2}{c}{} & \multicolumn{2}{c|}{} & \multicolumn{2}{c}{} & \multicolumn{2}{c}{} \\
		\multicolumn{2}{c|}{\multirow{2}[0]{*}{YOLOV5 \cite{YOLOV5}}} & \multicolumn{2}{c|}{\multirow{2}[0]{*}{RGB}} & \multicolumn{2}{c}{\multirow{2}[0]{*}{76.2 }} & \multicolumn{2}{c}{\multirow{2}[0]{*}{48.9 }} & \multicolumn{2}{c}{\multirow{2}[0]{*}{35.5 }} & \multicolumn{2}{c}{\multirow{2}[0]{*}{68.9 }} & \multicolumn{2}{c|}{\multirow{2}[0]{*}{40.4 }} & \multicolumn{2}{c}{\multirow{2}[0]{*}{54.0 }} & \multicolumn{2}{c}{\multirow{2}[0]{*}{30.2 }} \\
		\multicolumn{2}{c|}{} & \multicolumn{2}{c|}{} & \multicolumn{2}{c}{} & \multicolumn{2}{c}{} & \multicolumn{2}{c}{} & \multicolumn{2}{c}{} & \multicolumn{2}{c|}{} & \multicolumn{2}{c}{} & \multicolumn{2}{c}{} \\
		\multicolumn{2}{c|}{\multirow{2}[0]{*}{Oriented R-CNN \cite{orcnn}}} & \multicolumn{2}{c|}{\multirow{2}[0]{*}{RGB}} & \multicolumn{2}{c}{\multirow{2}[0]{*}{80.1 }} & \multicolumn{2}{c}{\multirow{2}[0]{*}{53.8 }} & \multicolumn{2}{c}{\multirow{2}[0]{*}{41.6 }} & \multicolumn{2}{c}{\multirow{2}[0]{*}{85.4 }} & \multicolumn{2}{c|}{\multirow{2}[0]{*}{43.3 }} & \multicolumn{2}{c}{\multirow{2}[0]{*}{60.8 }} & \multicolumn{2}{c}{\multirow{2}[0]{*}{32.7 }} \\
		\multicolumn{2}{c|}{} & \multicolumn{2}{c|}{} & \multicolumn{2}{c}{} & \multicolumn{2}{c}{} & \multicolumn{2}{c}{} & \multicolumn{2}{c}{} & \multicolumn{2}{c|}{} & \multicolumn{2}{c}{} & \multicolumn{2}{c}{} \\
		\multicolumn{2}{c|}{\multirow{2}[0]{*}{PKINet-S \cite{PKINet-S}}} & \multicolumn{2}{c|}{\multirow{2}[0]{*}{RGB}} & \multicolumn{2}{c}{\multirow{2}[0]{*}{76.4 }} & \multicolumn{2}{c}{\multirow{2}[0]{*}{53.0 }} & \multicolumn{2}{c}{\multirow{2}[0]{*}{40.8 }} & \multicolumn{2}{c}{\multirow{2}[0]{*}{79.1 }} & \multicolumn{2}{c|}{\multirow{2}[0]{*}{45.7 }} & \multicolumn{2}{c}{\multirow{2}[0]{*}{59.0 }} & \multicolumn{2}{c}{\multirow{2}[0]{*}{31.2 }} \\
		\multicolumn{2}{c|}{} & \multicolumn{2}{c|}{} & \multicolumn{2}{c}{} & \multicolumn{2}{c}{} & \multicolumn{2}{c}{} & \multicolumn{2}{c}{} & \multicolumn{2}{c|}{} & \multicolumn{2}{c}{} & \multicolumn{2}{c}{} \\
		\multicolumn{2}{c|}{\multirow{2}[0]{*}{Ours}} & \multicolumn{2}{c|}{\multirow{2}[0]{*}{RGB+Infrared}} & \multicolumn{2}{c}{\multirow{2}[0]{*}{\textbf{90.5} }} & \multicolumn{2}{c}{\multirow{2}[0]{*}{\textbf{79.4} }} & \multicolumn{2}{c}{\multirow{2}[0]{*}{\textbf{66.6} }} & \multicolumn{2}{c}{\multirow{2}[0]{*}{\textbf{90.0} }} & \multicolumn{2}{c|}{\multirow{2}[0]{*}{\textbf{67.8} }} & \multicolumn{2}{c}{\multirow{2}[0]{*}{\textbf{78.9} }} & \multicolumn{2}{c}{\multirow{2}[0]{*}{\textbf{51.2} }} \\
		\multicolumn{2}{c|}{} & \multicolumn{2}{c|}{} & \multicolumn{2}{c}{} & \multicolumn{2}{c}{} & \multicolumn{2}{c}{} & \multicolumn{2}{c}{} & \multicolumn{2}{c|}{} & \multicolumn{2}{c}{} & \multicolumn{2}{c}{} \\
		\hline
		\multicolumn{2}{c|}{\multirow{2}[1]{*}{RetinaNet \cite{retinanet}}} & \multicolumn{2}{c|}{\multirow{2}[1]{*}{Infrared}} & \multicolumn{2}{c}{\multirow{2}[1]{*}{88.8 }} & \multicolumn{2}{c}{\multirow{2}[1]{*}{35.4 }} & \multicolumn{2}{c}{\multirow{2}[1]{*}{39.5 }} & \multicolumn{2}{c}{\multirow{2}[1]{*}{76.5 }} & \multicolumn{2}{c|}{\multirow{2}[1]{*}{32.1 }} & \multicolumn{2}{c}{\multirow{2}[1]{*}{54.5 }} & \multicolumn{2}{c}{\multirow{2}[1]{*}{30.4 }} \\
		\multicolumn{2}{c|}{} & \multicolumn{2}{c|}{} & \multicolumn{2}{c}{} & \multicolumn{2}{c}{} & \multicolumn{2}{c}{} & \multicolumn{2}{c}{} & \multicolumn{2}{c|}{} & \multicolumn{2}{c}{} & \multicolumn{2}{c}{} \\
		\multicolumn{2}{c|}{\multirow{2}[0]{*}{$\text{R}^3\text{Det}$ \cite{r3det}}} & \multicolumn{2}{c|}{\multirow{2}[0]{*}{Infrared}} & \multicolumn{2}{c}{\multirow{2}[0]{*}{89.5 }} & \multicolumn{2}{c}{\multirow{2}[0]{*}{48.3 }} & \multicolumn{2}{c}{\multirow{2}[0]{*}{16.6 }} & \multicolumn{2}{c}{\multirow{2}[0]{*}{87.1 }} & \multicolumn{2}{c|}{\multirow{2}[0]{*}{39.9 }} & \multicolumn{2}{c}{\multirow{2}[0]{*}{62.3 }} & \multicolumn{2}{c}{\multirow{2}[0]{*}{36.7 }} \\
		\multicolumn{2}{c|}{} & \multicolumn{2}{c|}{} & \multicolumn{2}{c}{} & \multicolumn{2}{c}{} & \multicolumn{2}{c}{} & \multicolumn{2}{c}{} & \multicolumn{2}{c|}{} & \multicolumn{2}{c}{} & \multicolumn{2}{c}{} \\
		\multicolumn{2}{c|}{\multirow{2}[0]{*}{$\text{S}^2\text{ANet}$ \cite{S2ANet}}} & \multicolumn{2}{c|}{\multirow{2}[0]{*}{Infrared}} & \multicolumn{2}{c}{\multirow{2}[0]{*}{89.9 }} & \multicolumn{2}{c}{\multirow{2}[0]{*}{54.5 }} & \multicolumn{2}{c}{\multirow{2}[0]{*}{55.8 }} & \multicolumn{2}{c}{\multirow{2}[0]{*}{88.9 }} & \multicolumn{2}{c|}{\multirow{2}[0]{*}{48.4 }} & \multicolumn{2}{c}{\multirow{2}[0]{*}{67.5 }} & \multicolumn{2}{c}{\multirow{2}[0]{*}{40.4 }} \\
		\multicolumn{2}{c|}{} & \multicolumn{2}{c|}{} & \multicolumn{2}{c}{} & \multicolumn{2}{c}{} & \multicolumn{2}{c}{} & \multicolumn{2}{c}{} & \multicolumn{2}{c|}{} & \multicolumn{2}{c}{} & \multicolumn{2}{c}{} \\
		\multicolumn{2}{c|}{\multirow{2}[0]{*}{KFIoU \cite{kfiou}}} & \multicolumn{2}{c|}{\multirow{2}[0]{*}{Infrared}} & \multicolumn{2}{c}{\multirow{2}[0]{*}{90.2 }} & \multicolumn{2}{c}{\multirow{2}[0]{*}{65.0 }} & \multicolumn{2}{c}{\multirow{2}[0]{*}{48.5 }} & \multicolumn{2}{c}{\multirow{2}[0]{*}{89.0 }} & \multicolumn{2}{c|}{\multirow{2}[0]{*}{43.0 }} & \multicolumn{2}{c}{\multirow{2}[0]{*}{67.1 }} & \multicolumn{2}{c}{\multirow{2}[0]{*}{41.2 }} \\
		\multicolumn{2}{c|}{} & \multicolumn{2}{c|}{} & \multicolumn{2}{c}{} & \multicolumn{2}{c}{} & \multicolumn{2}{c}{} & \multicolumn{2}{c}{} & \multicolumn{2}{c|}{} & \multicolumn{2}{c}{} & \multicolumn{2}{c}{} \\
		\multicolumn{2}{c|}{\multirow{2}[0]{*}{ReDet \cite{redet}}} & \multicolumn{2}{c|}{\multirow{2}[0]{*}{Infrared}} & \multicolumn{2}{c}{\multirow{2}[0]{*}{89.8 }} & \multicolumn{2}{c}{\multirow{2}[0]{*}{53.9 }} & \multicolumn{2}{c}{\multirow{2}[0]{*}{43.3 }} & \multicolumn{2}{c}{\multirow{2}[0]{*}{84.8 }} & \multicolumn{2}{c|}{\multirow{2}[0]{*}{33.2 }} & \multicolumn{2}{c}{\multirow{2}[0]{*}{61.0 }} & \multicolumn{2}{c}{\multirow{2}[0]{*}{36.2 }} \\
		\multicolumn{2}{c|}{} & \multicolumn{2}{c|}{} & \multicolumn{2}{c}{} & \multicolumn{2}{c}{} & \multicolumn{2}{c}{} & \multicolumn{2}{c}{} & \multicolumn{2}{c|}{} & \multicolumn{2}{c}{} & \multicolumn{2}{c}{} \\
		\multicolumn{2}{c|}{\multirow{2}[0]{*}{GWD \cite{GWD}}} & \multicolumn{2}{c|}{\multirow{2}[0]{*}{Infrared}} & \multicolumn{2}{c}{\multirow{2}[0]{*}{89.8 }} & \multicolumn{2}{c}{\multirow{2}[0]{*}{39.6 }} & \multicolumn{2}{c}{\multirow{2}[0]{*}{26.5 }} & \multicolumn{2}{c}{\multirow{2}[0]{*}{75.0 }} & \multicolumn{2}{c|}{\multirow{2}[0]{*}{23.2 }} & \multicolumn{2}{c}{\multirow{2}[0]{*}{50.8 }} & \multicolumn{2}{c}{\multirow{2}[0]{*}{30.7 }} \\
		\multicolumn{2}{c|}{} & \multicolumn{2}{c|}{} & \multicolumn{2}{c}{} & \multicolumn{2}{c}{} & \multicolumn{2}{c}{} & \multicolumn{2}{c}{} & \multicolumn{2}{c|}{} & \multicolumn{2}{c}{} & \multicolumn{2}{c}{} \\
		\multicolumn{2}{c|}{\multirow{2}[0]{*}{PKINet-S \cite{PKINet-S}}} & \multicolumn{2}{c|}{\multirow{2}[0]{*}{Infrared}} & \multicolumn{2}{c}{\multirow{2}[0]{*}{90.2 }} & \multicolumn{2}{c}{\multirow{2}[0]{*}{67.3 }} & \multicolumn{2}{c}{\multirow{2}[0]{*}{56.3 }} & \multicolumn{2}{c}{\multirow{2}[0]{*}{88.8 }} & \multicolumn{2}{c|}{\multirow{2}[0]{*}{51.2 }} & \multicolumn{2}{c}{\multirow{2}[0]{*}{70.8 }} & \multicolumn{2}{c}{\multirow{2}[0]{*}{43.8 }} \\
		\multicolumn{2}{c|}{} & \multicolumn{2}{c|}{} & \multicolumn{2}{c}{} & \multicolumn{2}{c}{} & \multicolumn{2}{c}{} & \multicolumn{2}{c}{} & \multicolumn{2}{c|}{} & \multicolumn{2}{c}{} & \multicolumn{2}{c}{} \\
		\multicolumn{2}{c|}{\multirow{2}[0]{*}{DTNet \cite{dtnet}}} & \multicolumn{2}{c|}{\multirow{2}[0]{*}{Infrared}} & \multicolumn{2}{c}{\multirow{2}[0]{*}{90.2 }} & \multicolumn{2}{c}{\multirow{2}[0]{*}{65.7 }} & \multicolumn{2}{c}{\multirow{2}[0]{*}{\textbf{78.1} }} & \multicolumn{2}{c}{\multirow{2}[0]{*}{89.2 }} & \multicolumn{2}{c|}{\multirow{2}[0]{*}{\textbf{67.9} }} & \multicolumn{2}{c}{\multirow{2}[0]{*}{78.2 }} & \multicolumn{2}{c}{\multirow{2}[0]{*}{\textbf{52.9} }} \\
		\multicolumn{2}{c|}{} & \multicolumn{2}{c|}{} & \multicolumn{2}{c}{} & \multicolumn{2}{c}{} & \multicolumn{2}{c}{} & \multicolumn{2}{c}{} & \multicolumn{2}{c|}{} & \multicolumn{2}{c}{} & \multicolumn{2}{c}{} \\
		\multicolumn{2}{c|}{\multirow{2}[0]{*}{Ours}} & \multicolumn{2}{c|}{\multirow{2}[0]{*}{RGB+Infrared}} & \multicolumn{2}{c}{\multirow{2}[0]{*}{\textbf{90.5} }} & \multicolumn{2}{c}{\multirow{2}[0]{*}{\textbf{79.4} }} & \multicolumn{2}{c}{\multirow{2}[0]{*}{66.6 }} & \multicolumn{2}{c}{\multirow{2}[0]{*}{\textbf{90.0} }} & \multicolumn{2}{c|}{\multirow{2}[0]{*}{67.8 }} & \multicolumn{2}{c}{\multirow{2}[0]{*}{\textbf{78.9} }} & \multicolumn{2}{c}{\multirow{2}[0]{*}{51.2 }} \\
		\multicolumn{2}{c|}{} & \multicolumn{2}{c|}{} & \multicolumn{2}{c}{} & \multicolumn{2}{c}{} & \multicolumn{2}{c}{} & \multicolumn{2}{c}{} & \multicolumn{2}{c|}{} & \multicolumn{2}{c}{} & \multicolumn{2}{c}{} \\
		\hline
		\multicolumn{2}{c|}{\multirow{2}[1]{*}{UA-CMDet \cite{dronevehicle}}} & \multicolumn{2}{c|}{\multirow{2}[1]{*}{RGB+Infrared}} & \multicolumn{2}{c}{\multirow{2}[1]{*}{87.5 }} & \multicolumn{2}{c}{\multirow{2}[1]{*}{60.7 }} & \multicolumn{2}{c}{\multirow{2}[1]{*}{46.8 }} & \multicolumn{2}{c}{\multirow{2}[1]{*}{87.1 }} & \multicolumn{2}{c|}{\multirow{2}[1]{*}{38.0 }} & \multicolumn{2}{c}{\multirow{2}[1]{*}{64.0 }} & \multicolumn{2}{c}{\multirow{2}[1]{*}{40.1 }} \\
		\multicolumn{2}{c|}{} & \multicolumn{2}{c|}{} & \multicolumn{2}{c}{} & \multicolumn{2}{c}{} & \multicolumn{2}{c}{} & \multicolumn{2}{c}{} & \multicolumn{2}{c|}{} & \multicolumn{2}{c}{} & \multicolumn{2}{c}{} \\
		\multicolumn{2}{c|}{\multirow{2}[0]{*}
		{GLFNet \cite{GLFNet}}} & \multicolumn{2}{c|}{\multirow{2}[0]{*}{RGB+Infrared}} & \multicolumn{2}{c}{\multirow{2}[0]{*}{90.3 }} & \multicolumn{2}{c}{\multirow{2}[0]{*}{72.7 }} & \multicolumn{2}{c}{\multirow{2}[0]{*}{53.6 }} & \multicolumn{2}{c}{\multirow{2}[0]{*}{88.0 }} & \multicolumn{2}{c|}{\multirow{2}[0]{*}{52.6 }} & \multicolumn{2}{c}{\multirow{2}[0]{*}{71.4 }} & \multicolumn{2}{c}{\multirow{2}[0]{*}{42.9 }} \\
		\multicolumn{2}{c|}{} & \multicolumn{2}{c|}{} & \multicolumn{2}{c}{} & \multicolumn{2}{c}{} & \multicolumn{2}{c}{} & \multicolumn{2}{c}{} & \multicolumn{2}{c|}{} & \multicolumn{2}{c}{} & \multicolumn{2}{c}{} \\
		\multicolumn{2}{c|}{\multirow{2}[0]{*}{AR-CNN \cite{AR-CNN}}} & \multicolumn{2}{c|}{\multirow{2}[0]{*}{RGB+Infrared}} & \multicolumn{2}{c}{\multirow{2}[0]{*}{90.1 }} & \multicolumn{2}{c}{\multirow{2}[0]{*}{64.8 }} & \multicolumn{2}{c}{\multirow{2}[0]{*}{62.1 }} & \multicolumn{2}{c}{\multirow{2}[0]{*}{89.4 }} & \multicolumn{2}{c|}{\multirow{2}[0]{*}{51.5 }} & \multicolumn{2}{c}{\multirow{2}[0]{*}{71.6 }} & \multicolumn{2}{c}{\multirow{2}[0]{*}{-}} \\
		\multicolumn{2}{c|}{} & \multicolumn{2}{c|}{} & \multicolumn{2}{c}{} & \multicolumn{2}{c}{} & \multicolumn{2}{c}{} & \multicolumn{2}{c}{} & \multicolumn{2}{c|}{} & \multicolumn{2}{c}{} & \multicolumn{2}{c}{} \\
		\multicolumn{2}{c|}{\multirow{2}[0]{*}{MBNet \cite{MBNet}}} & \multicolumn{2}{c|}{\multirow{2}[0]{*}{RGB+Infrared}} & \multicolumn{2}{c}{\multirow{2}[0]{*}{90.1 }} & \multicolumn{2}{c}{\multirow{2}[0]{*}{64.4 }} & \multicolumn{2}{c}{\multirow{2}[0]{*}{62.4 }} & \multicolumn{2}{c}{\multirow{2}[0]{*}{88.8 }} & \multicolumn{2}{c|}{\multirow{2}[0]{*}{53.6 }} & \multicolumn{2}{c}{\multirow{2}[0]{*}{71.9 }} & \multicolumn{2}{c}{\multirow{2}[0]{*}{-}} \\
		\multicolumn{2}{c|}{} & \multicolumn{2}{c|}{} & \multicolumn{2}{c}{} & \multicolumn{2}{c}{} & \multicolumn{2}{c}{} & \multicolumn{2}{c}{} & \multicolumn{2}{c|}{} & \multicolumn{2}{c}{} & \multicolumn{2}{c}{} \\
		\multicolumn{2}{c|}{\multirow{2}[0]{*}{TSFADet \cite{tsr}}} & \multicolumn{2}{c|}{\multirow{2}[0]{*}{RGB+Infrared}} & \multicolumn{2}{c}{\multirow{2}[0]{*}{89.9 }} & \multicolumn{2}{c}{\multirow{2}[0]{*}{67.9 }} & \multicolumn{2}{c}{\multirow{2}[0]{*}{63.7 }} & \multicolumn{2}{c}{\multirow{2}[0]{*}{89.8 }} & \multicolumn{2}{c|}{\multirow{2}[0]{*}{54.0 }} & \multicolumn{2}{c}{\multirow{2}[0]{*}{73.1 }} & \multicolumn{2}{c}{\multirow{2}[0]{*}{-}} \\
		\multicolumn{2}{c|}{} & \multicolumn{2}{c|}{} & \multicolumn{2}{c}{} & \multicolumn{2}{c}{} & \multicolumn{2}{c}{} & \multicolumn{2}{c}{} & \multicolumn{2}{c|}{} & \multicolumn{2}{c}{} & \multicolumn{2}{c}{} \\
		\multicolumn{2}{c|}{\multirow{2}[0]{*}{CIAN \cite{CIAN}}} & \multicolumn{2}{c|}{\multirow{2}[0]{*}{RGB+Infrared}} & \multicolumn{2}{c}{\multirow{2}[0]{*}{90.1 }} & \multicolumn{2}{c}{\multirow{2}[0]{*}{63.8 }} & \multicolumn{2}{c}{\multirow{2}[0]{*}{60.7 }} & \multicolumn{2}{c}{\multirow{2}[0]{*}{89.1 }} & \multicolumn{2}{c|}{\multirow{2}[0]{*}{50.3 }} & \multicolumn{2}{c}{\multirow{2}[0]{*}{70.8 }} & \multicolumn{2}{c}{\multirow{2}[0]{*}{-}} \\
		\multicolumn{2}{c|}{} & \multicolumn{2}{c|}{} & \multicolumn{2}{c}{} & \multicolumn{2}{c}{} & \multicolumn{2}{c}{} & \multicolumn{2}{c}{} & \multicolumn{2}{c|}{} & \multicolumn{2}{c}{} & \multicolumn{2}{c}{} \\
		\multicolumn{2}{c|}{\multirow{2}[0]{*}{$\text{C}^2\text{Former}$ \cite{c2former}}} & \multicolumn{2}{c|}{\multirow{2}[0]{*}{RGB+Infrared}} & \multicolumn{2}{c}{\multirow{2}[0]{*}{90.2 }} & \multicolumn{2}{c}{\multirow{2}[0]{*}{68.3 }} & \multicolumn{2}{c}{\multirow{2}[0]{*}{64.4 }} & \multicolumn{2}{c}{\multirow{2}[0]{*}{89.8 }} & \multicolumn{2}{c|}{\multirow{2}[0]{*}{58.5 }} & \multicolumn{2}{c}{\multirow{2}[0]{*}{74.2 }} & \multicolumn{2}{c}{\multirow{2}[0]{*}{47.5 }} \\
		\multicolumn{2}{c|}{} & \multicolumn{2}{c|}{} & \multicolumn{2}{c}{} & \multicolumn{2}{c}{} & \multicolumn{2}{c}{} & \multicolumn{2}{c}{} & \multicolumn{2}{c|}{} & \multicolumn{2}{c}{} & \multicolumn{2}{c}{} \\
		\multicolumn{2}{c|}{\multirow{2}[0]{*}{TarDAL \cite{tardal}}} & \multicolumn{2}{c|}{\multirow{2}[0]{*}{RGB+Infrared}} & \multicolumn{2}{c}{\multirow{2}[0]{*}{89.5 }} & \multicolumn{2}{c}{\multirow{2}[0]{*}{68.3 }} & \multicolumn{2}{c}{\multirow{2}[0]{*}{56.1 }} & \multicolumn{2}{c}{\multirow{2}[0]{*}{89.4 }} & \multicolumn{2}{c|}{\multirow{2}[0]{*}{59.3 }} & \multicolumn{2}{c}{\multirow{2}[0]{*}{72.6 }} & \multicolumn{2}{c}{\multirow{2}[0]{*}{43.3 }} \\
		\multicolumn{2}{c|}{} & \multicolumn{2}{c|}{} & \multicolumn{2}{c}{} & \multicolumn{2}{c}{} & \multicolumn{2}{c}{} & \multicolumn{2}{c}{} & \multicolumn{2}{c|}{} & \multicolumn{2}{c}{} & \multicolumn{2}{c}{} \\
		\multicolumn{2}{c|}{\multirow{2}[0]{*}{TFDet \cite{TFDet}}} & \multicolumn{2}{c|}{\multirow{2}[0]{*}{RGB+Infrared}} & \multicolumn{2}{c}{\multirow{2}[0]{*}{90.1 }} & \multicolumn{2}{c}{\multirow{2}[0]{*}{63.4 }} & \multicolumn{2}{c}{\multirow{2}[0]{*}{48.0 }} & \multicolumn{2}{c}{\multirow{2}[0]{*}{87.7 }} & \multicolumn{2}{c|}{\multirow{2}[0]{*}{48.9 }} & \multicolumn{2}{c}{\multirow{2}[0]{*}{67.6 }} & \multicolumn{2}{c}{\multirow{2}[0]{*}{- }} \\
		\multicolumn{2}{c|}{} & \multicolumn{2}{c|}{} & \multicolumn{2}{c}{} & \multicolumn{2}{c}{} & \multicolumn{2}{c}{} & \multicolumn{2}{c}{} & \multicolumn{2}{c|}{} & \multicolumn{2}{c}{} & \multicolumn{2}{c}{} \\
		\multicolumn{2}{c|}{\multirow{2}[0]{*}
		{DMM \cite{dmm}}} & \multicolumn{2}{c|}{\multirow{2}[0]{*}{RGB+Infrared}} & \multicolumn{2}{c}{\multirow{2}[0]{*}{90.4 }} & \multicolumn{2}{c}{\multirow{2}[0]{*}{78.9 }} & \multicolumn{2}{c}{\multirow{2}[0]{*}{66.2 }} & \multicolumn{2}{c}{\multirow{2}[0]{*}{89.5 }} & \multicolumn{2}{c|}{\multirow{2}[0]{*}{\textbf{68.3} }} & \multicolumn{2}{c}{\multirow{2}[0]{*}{78.6 }} & \multicolumn{2}{c}{\multirow{2}[0]{*}{50.6 }} \\
		\multicolumn{2}{c|}{} & \multicolumn{2}{c|}{} & \multicolumn{2}{c}{} & \multicolumn{2}{c}{} & \multicolumn{2}{c}{} & \multicolumn{2}{c}{} & \multicolumn{2}{c|}{} & \multicolumn{2}{c}{} & \multicolumn{2}{c}{} \\
		\multicolumn{2}{c|}{\multirow{2}[0]{*}{AFFNet \cite{AFFNet}}} & \multicolumn{2}{c|}{\multirow{2}[0]{*}{RGB+Infrared}} & \multicolumn{2}{c}{\multirow{2}[0]{*}{90.2 }} & \multicolumn{2}{c}{\multirow{2}[0]{*}{66.0 }} & \multicolumn{2}{c}{\multirow{2}[0]{*}{50.9 }} & \multicolumn{2}{c}{\multirow{2}[0]{*}{88.1 }} & \multicolumn{2}{c|}{\multirow{2}[0]{*}{48.8 }} & \multicolumn{2}{c}{\multirow{2}[0]{*}{68.8 }} & \multicolumn{2}{c}{\multirow{2}[0]{*}{- }} \\
		\multicolumn{2}{c|}{} & \multicolumn{2}{c|}{} & \multicolumn{2}{c}{} & \multicolumn{2}{c}{} & \multicolumn{2}{c}{} & \multicolumn{2}{c}{} & \multicolumn{2}{c|}{} & \multicolumn{2}{c}{} & \multicolumn{2}{c}{} \\
		\multicolumn{2}{c|}{\multirow{2}[0]{*}{Ours}} & \multicolumn{2}{c|}{\multirow{2}[0]{*}{RGB+Infrared}} & \multicolumn{2}{c}{\multirow{2}[0]{*}{\textbf{90.5} }} & \multicolumn{2}{c}{\multirow{2}[0]{*}{\textbf{79.4} }} & \multicolumn{2}{c}{\multirow{2}[0]{*}{\textbf{66.6} }} & \multicolumn{2}{c}{\multirow{2}[0]{*}{\textbf{90.0} }} & \multicolumn{2}{c|}{\multirow{2}[0]{*}{67.8 }} & \multicolumn{2}{c}{\multirow{2}[0]{*}{\textbf{78.9} }} & \multicolumn{2}{c}{\multirow{2}[0]{*}{\textbf{51.2} }} \\
		\multicolumn{2}{c|}{} & \multicolumn{2}{c|}{} & \multicolumn{2}{c}{} & \multicolumn{2}{c}{} & \multicolumn{2}{c}{} & \multicolumn{2}{c}{} & \multicolumn{2}{c|}{} & \multicolumn{2}{c}{} & \multicolumn{2}{c}{} \\
		\hline
	\end{tabular}}
\end{table*}%

\section{Experiment}
\subsection{Datasets and Metrics}
DEPFusion  is evaluated on DroneVehicle \cite{dronevehicle} and VEDAI \cite{vedai} benchmarks.

\subsubsection{DroneVehicle Dataset}
The DroneVehicle dataset is a large RGB-Infrared remote sensing dataset obtained by the real time flying of drones. The scenarios includes parking lots, roads, residential areas and etc. in daylight and nighttime. It contains 28,439 pairs of RGB and IR images with five classes of instances, including cars, trucks, freight cars, vans and buses. All of the classes have 95,3087 annotated bboxes and labels. The images are sized
at 840 $\times$ 712 pixels with an additional 100 pixels white
border and the actual size is 640 $\times $ 512  pixels. 

\subsubsection{VEDAI Dataset}
VEDAI dataset provides 1,246 paired RGB and infrared aerial images with sizes of 1024 $\times $ 1024 or 512 $\times$ 512 pixels for vehicle detection in varied urban/rural environments. Its oriented annotations cover 9 categories: core vehicle types such as cars, trucks, vans etc. and auxiliary objects such as airplanes, boats etc.

\subsubsection{Metrics}
The metrics we choose is mean Average Precision (mAP) which includes mAP@0.5 and mAP@0.5:0.95. mAP@0.5 represents that predicted bounding box is regarded as correct if the
Intersection over Union (IoU) with the ground truth is at
least 0.5. mAP@0.5:0.95 is the mean value of mAP under different IoU thresholds, whose range is from 0.5 to 0.95 with step size 0.05.

\subsection{Implementation Details}
DEPFusion  is built on MMdetection \cite{mmdetection} and MMrotate \cite{mmrotate} platform. All of the experiments are conducted on four NVIDIA RTX 3090 GPUs with 24GB of memory. The environment of training and testing is Pytorch 2.1.2 and CUDA 11.8. DEPFusion  is trained with 12 epochs and batch size of 4. The AdamW optimizer is adopted with an initial learning rate of 0.0001 and a weight decay of 0.05. We adopt Vision-Mamba as our backbone network and $\text{S}^2\text{ANet}$ as our detection head.  For the settings of CSWM block, Haar is chosen as wavelet basis, and the level of 2D-DWT is set to 2.

The sizes of input images are 512 $\times$ 640 for DroneVehicle dataset and  512 $\times$ 512 for VEDAI dataset respectively.
Following official setting \cite{dronevehicle,vedai}, the DroneVehicle dataset are divided into 17990, 1469 and 8980 pairs of RGB-IR image as train, validation and test set, while the VEDAI dataset are divided into 971, 100, 175  pairs of RGB-IR image as train validation and test set. All of the groundtruth is annotated with Orientated Bounding Box (OBB) format.

\begin{figure*}[t]
	\centering
	\includegraphics[width=7in]{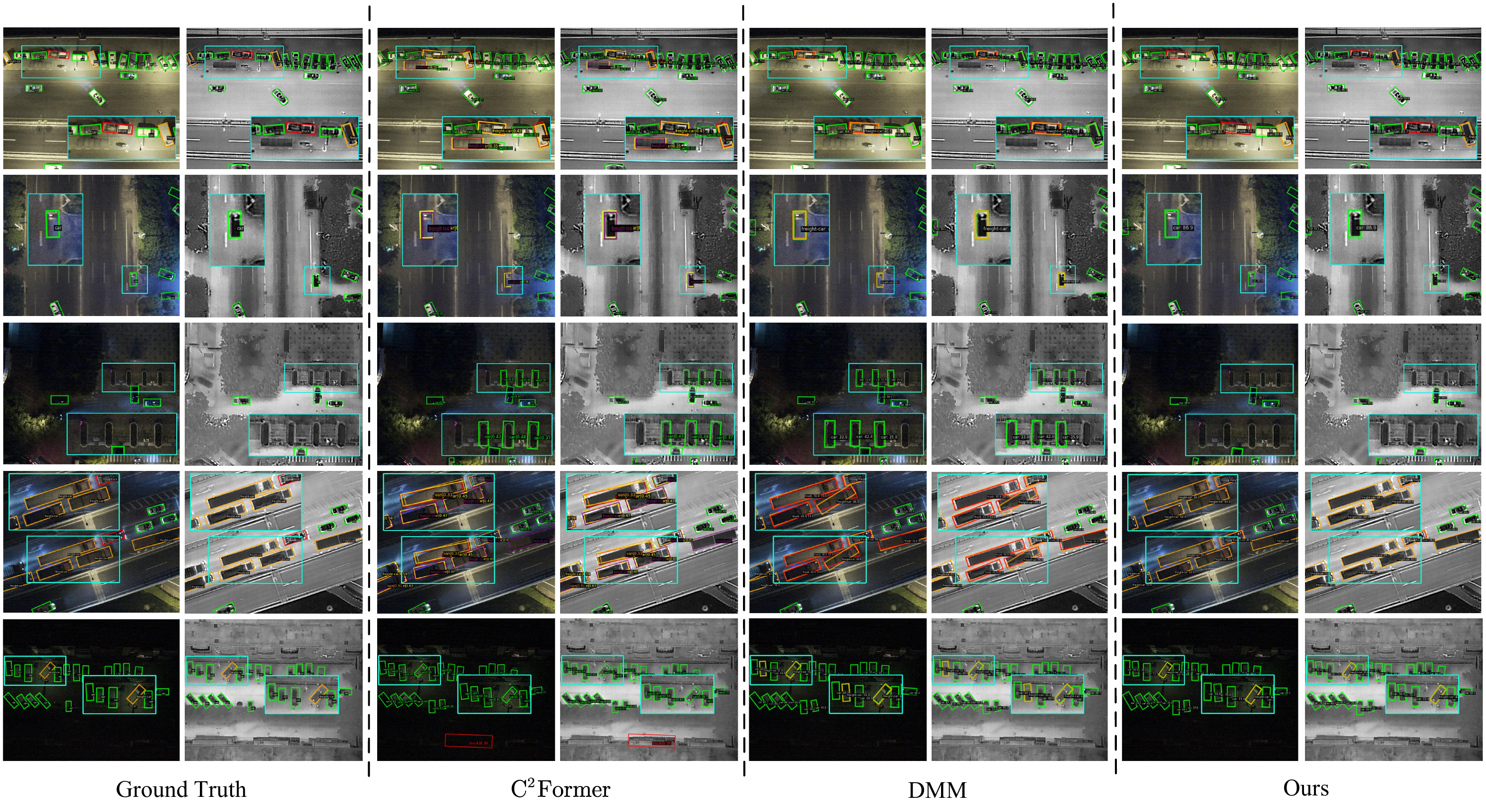}
	\caption{The visual comparison results on the DroneVehicle datasets. The advantages of our method are marked with blue circles. DEPFusion  overcomes incorrect classification, missed detections, and false detections in the low-light scenarios.	\label{showdv}
	}
\end{figure*}

\begin{table}[t]
	\centering
	\small
	\caption{The complexity comparison between our DEPFusion  and SOTA on NVIDIA RTX 3090 GPU.\label{complex}}
	\renewcommand{\arraystretch}{0.7}
	\setlength{\tabcolsep}{0.5mm}{
		\begin{tabular}{cc|c|c|cccccc}
			\hline
			\multicolumn{2}{c|}{\multirow{2}[1]{*}{Methods}} & \multicolumn{2}{c|}{\multirow{2}[1]{*}{Input Size}} & \multicolumn{2}{c}{\multirow{2}[1]{*}{Para (MB) $\downarrow$}} & \multicolumn{2}{c}{\multirow{2}[1]{*}{FLOPs (G) $\downarrow$}} & \multicolumn{2}{c}{\multirow{2}[1]{*}{mAP@0.5 $\uparrow$}} \\
			\multicolumn{2}{c|}{} & \multicolumn{2}{c|}{} & \multicolumn{2}{c}{} & \multicolumn{2}{c}{} & \multicolumn{2}{c}{} \\
			\hline
			\multicolumn{2}{c|}{\multirow{2}[1]{*}{TSFADet}} & \multicolumn{2}{c|}{\multirow{8}[1]{*}{512 $\times$ 640}} & \multicolumn{2}{c}{\multirow{2}[1]{*}{104.7}} & \multicolumn{2}{c}{\multirow{2}[1]{*}{109.8}} & \multicolumn{2}{c}{\multirow{2}[1]{*}{73.1}} \\
			\multicolumn{2}{c|}{} & \multicolumn{2}{c|}{} & \multicolumn{2}{c}{} & \multicolumn{2}{c}{} & \multicolumn{2}{c}{} \\
			\multicolumn{2}{c|}{\multirow{2}[0]{*}{$\text{C}^2\text{Former}$}} & \multicolumn{2}{c|}{} & \multicolumn{2}{c}{\multirow{2}[0]{*}{118.5}} & \multicolumn{2}{c}{\multirow{2}[0]{*}{115.6}} & \multicolumn{2}{c}{\multirow{2}[0]{*}{74.2}} \\
			\multicolumn{2}{c|}{} & \multicolumn{2}{c|}{} & \multicolumn{2}{c}{} & \multicolumn{2}{c}{} & \multicolumn{2}{c}{} \\
			\multicolumn{2}{c|}{\multirow{2}[0]{*}{DMM}} & \multicolumn{2}{c|}{} & \multicolumn{2}{c}{\multirow{2}[0]{*}{88.1}} & \multicolumn{2}{c}{\multirow{2}[0]{*}{108.8}} & \multicolumn{2}{c}{\multirow{2}[0]{*}{78.6}} \\
			\multicolumn{2}{c|}{} & \multicolumn{2}{c|}{} & \multicolumn{2}{c}{} & \multicolumn{2}{c}{} & \multicolumn{2}{c}{} \\
			\multicolumn{2}{c|}{\multirow{2}[0]{*}{Ours}} & \multicolumn{2}{c|}{} & \multicolumn{2}{c}{\multirow{2}[0]{*}{\textbf{80.3}}} & \multicolumn{2}{c}{\multirow{2}[0]{*}{\textbf{108.3}}} & \multicolumn{2}{c}{\multirow{2}[0]{*}{\textbf{78.9}}} \\
			\multicolumn{2}{c|}{} & \multicolumn{2}{c|}{} & \multicolumn{2}{c}{} & \multicolumn{2}{c}{} & \multicolumn{2}{c}{} \\
			\hline
	\end{tabular}}
\end{table}%
\begin{figure*}[t]
	\centering
	\includegraphics[width=7in]{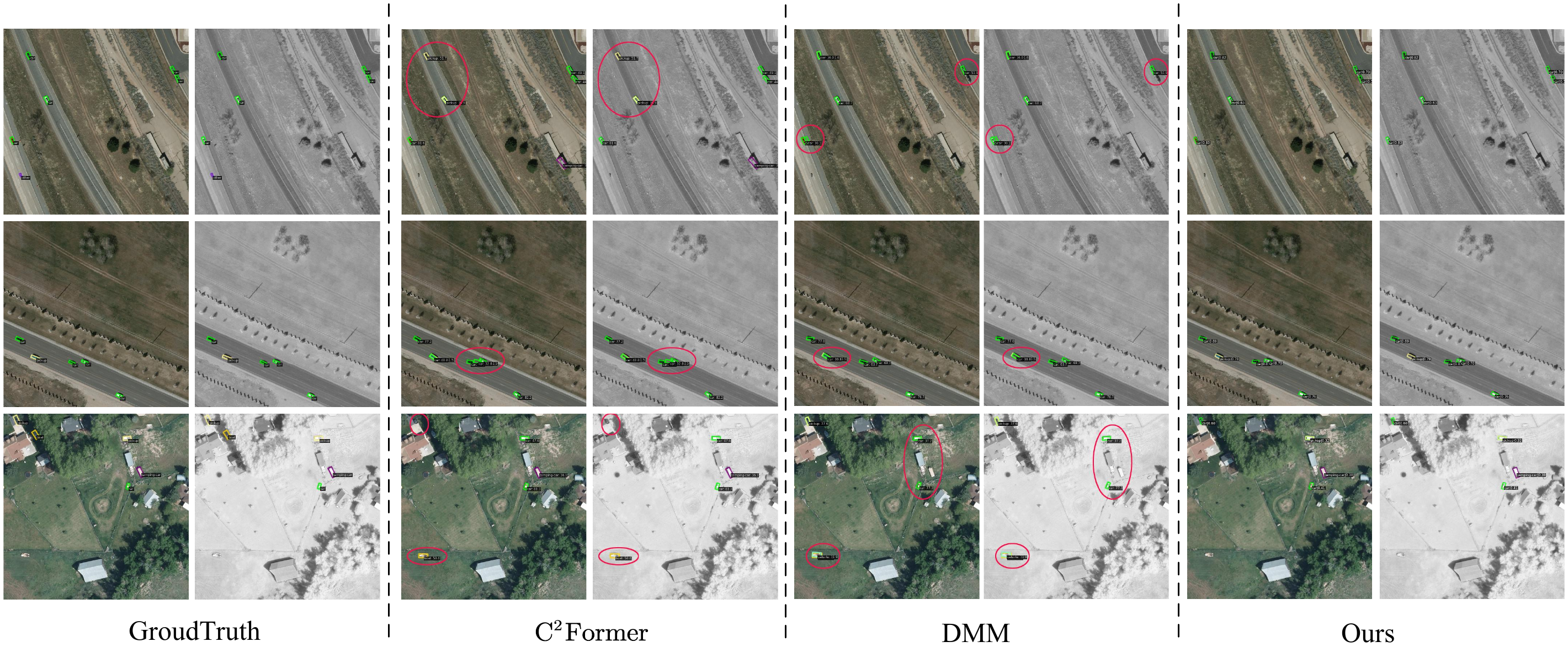}
	\caption{The visual comparison results on the VEDAI datasets. The weakness of SOTA methods are marked with red circles. \label{vedai}
	}
\end{figure*}
\begin{table*}[t]
	\centering
	\small
	\caption{Comparative experiment results on the VEDAI dataset, the best results are marked with bold font. \label{vedai1}}
	\renewcommand{\arraystretch}{0.7}
	\setlength{\tabcolsep}{0.7mm}{
		\begin{tabular}{cc|cc|cccccccccccccccccc|cccc}
			\hline
			\multicolumn{2}{c|}{\multirow{2}[1]{*}{Methods}} & \multicolumn{2}{c|}{\multirow{2}[1]{*}{Modality}} & \multicolumn{2}{c}{\multirow{2}[1]{*}{Car}} & \multicolumn{2}{c}{\multirow{2}[1]{*}{Truck}} & \multicolumn{2}{c}{\multirow{2}[1]{*}{Tractor}} & \multicolumn{2}{c}{\multirow{2}[1]{*}{Camping Car}} & \multicolumn{2}{c}{\multirow{2}[1]{*}{Van}} & \multicolumn{2}{c}{\multirow{2}[1]{*}{Pick-up}} & \multicolumn{2}{c}{\multirow{2}[1]{*}{Boat}} & \multicolumn{2}{c}{\multirow{2}[1]{*}{Plane}} & \multicolumn{2}{c|}{\multirow{2}[1]{*}{Others}} & \multicolumn{2}{c}{\multirow{2}[1]{*}{mAP50}} & \multicolumn{2}{c}{\multirow{2}[1]{*}{mAP0.5:0.95}} \\
			\multicolumn{2}{c|}{} & \multicolumn{2}{c|}{} & \multicolumn{2}{c}{} & \multicolumn{2}{c}{} & \multicolumn{2}{c}{} & \multicolumn{2}{c}{} & \multicolumn{2}{c}{} & \multicolumn{2}{c}{} & \multicolumn{2}{c}{} & \multicolumn{2}{c}{} & \multicolumn{2}{c|}{} & \multicolumn{2}{c}{} & \multicolumn{2}{c}{} \\
			\hline
			\multicolumn{2}{c|}{\multirow{2}[0]{*}{RetinaNet \cite{retinanet}}} & \multicolumn{2}{c|}{\multirow{2}[0]{*}{RGB}} & \multicolumn{2}{c}{\multirow{2}[0]{*}{48.9 }} & \multicolumn{2}{c}{\multirow{2}[0]{*}{16.8 }} & \multicolumn{2}{c}{\multirow{2}[0]{*}{15.9}} & \multicolumn{2}{c}{\multirow{2}[0]{*}{21.4}} & \multicolumn{2}{c}{\multirow{2}[0]{*}{5.9 }} & \multicolumn{2}{c}{\multirow{2}[0]{*}{37.5}} & \multicolumn{2}{c}{\multirow{2}[0]{*}{4.4 }} & \multicolumn{2}{c}{\multirow{2}[0]{*}{21.2}} & \multicolumn{2}{c|}{\multirow{2}[0]{*}{14.1 }} & \multicolumn{2}{c}{\multirow{2}[0]{*}{20.7 }} & \multicolumn{2}{c}{\multirow{2}[0]{*}{8.5 }} \\
			\multicolumn{2}{c|}{} & \multicolumn{2}{c|}{} & \multicolumn{2}{c}{} & \multicolumn{2}{c}{} & \multicolumn{2}{c}{} & \multicolumn{2}{c}{} & \multicolumn{2}{c}{} & \multicolumn{2}{c}{} & \multicolumn{2}{c}{} & \multicolumn{2}{c}{} & \multicolumn{2}{c|}{} & \multicolumn{2}{c}{} & \multicolumn{2}{c}{} \\
			\multicolumn{2}{c|}{\multirow{2}[0]{*}{YOLO-S \cite{yolos}}} & \multicolumn{2}{c|}{\multirow{2}[0]{*}{RGB}} & \multicolumn{2}{c}{\multirow{2}[0]{*}{\textbf{95.5 }}} & \multicolumn{2}{c}{\multirow{2}[0]{*}{65.5 }} & \multicolumn{2}{c}{\multirow{2}[0]{*}{63.8}} & \multicolumn{2}{c}{\multirow{2}[0]{*}{-}} & \multicolumn{2}{c}{\multirow{2}[0]{*}{70.7 }} & \multicolumn{2}{c}{\multirow{2}[0]{*}{-}} & \multicolumn{2}{c}{\multirow{2}[0]{*}{74.2 }} & \multicolumn{2}{c}{\multirow{2}[0]{*}{75}} & \multicolumn{2}{c|}{\multirow{2}[0]{*}{47.9 }} & \multicolumn{2}{c}{\multirow{2}[0]{*}{70.4 }} & \multicolumn{2}{c}{\multirow{2}[0]{*}{-}} \\
			\multicolumn{2}{c|}{} & \multicolumn{2}{c|}{} & \multicolumn{2}{c}{} & \multicolumn{2}{c}{} & \multicolumn{2}{c}{} & \multicolumn{2}{c}{} & \multicolumn{2}{c}{} & \multicolumn{2}{c}{} & \multicolumn{2}{c}{} & \multicolumn{2}{c}{} & \multicolumn{2}{c|}{} & \multicolumn{2}{c}{} & \multicolumn{2}{c}{} \\
			\multicolumn{2}{c|}{\multirow{2}[0]{*}{YOLOV3 \cite{yolov3}}} & \multicolumn{2}{c|}{\multirow{2}[0]{*}{RGB}} & \multicolumn{2}{c}{\multirow{2}[0]{*}{80.4 }} & \multicolumn{2}{c}{\multirow{2}[0]{*}{38.6 }} & \multicolumn{2}{c}{\multirow{2}[0]{*}{57.7}} & \multicolumn{2}{c}{\multirow{2}[0]{*}{66.5}} & \multicolumn{2}{c}{\multirow{2}[0]{*}{\textbf{76.9} }} & \multicolumn{2}{c}{\multirow{2}[0]{*}{67.8}} & \multicolumn{2}{c}{\multirow{2}[0]{*}{26.2 }} & \multicolumn{2}{c}{\multirow{2}[0]{*}{88.7}} & \multicolumn{2}{c|}{\multirow{2}[0]{*}{51.3 }} & \multicolumn{2}{c}{\multirow{2}[0]{*}{61.6 }} & \multicolumn{2}{c}{\multirow{2}[0]{*}{-}} \\
			\multicolumn{2}{c|}{} & \multicolumn{2}{c|}{} & \multicolumn{2}{c}{} & \multicolumn{2}{c}{} & \multicolumn{2}{c}{} & \multicolumn{2}{c}{} & \multicolumn{2}{c}{} & \multicolumn{2}{c}{} & \multicolumn{2}{c}{} & \multicolumn{2}{c}{} & \multicolumn{2}{c|}{} & \multicolumn{2}{c}{} & \multicolumn{2}{c}{} \\
			\multicolumn{2}{c|}{\multirow{2}[0]{*}{ROITransfomer \cite{roitrans}}} & \multicolumn{2}{c|}{\multirow{2}[0]{*}{RGB}} & \multicolumn{2}{c}{\multirow{2}[0]{*}{77.3 }} & \multicolumn{2}{c}{\multirow{2}[0]{*}{56.1 }} & \multicolumn{2}{c}{\multirow{2}[0]{*}{64.7}} & \multicolumn{2}{c}{\multirow{2}[0]{*}{73.6}} & \multicolumn{2}{c}{\multirow{2}[0]{*}{60.2 }} & \multicolumn{2}{c}{\multirow{2}[0]{*}{71.5}} & \multicolumn{2}{c}{\multirow{2}[0]{*}{56.7 }} & \multicolumn{2}{c}{\multirow{2}[0]{*}{85.7}} & \multicolumn{2}{c|}{\multirow{2}[0]{*}{42.8 }} & \multicolumn{2}{c}{\multirow{2}[0]{*}{65.4 }} & \multicolumn{2}{c}{\multirow{2}[0]{*}{27.6 }} \\
			\multicolumn{2}{c|}{} & \multicolumn{2}{c|}{} & \multicolumn{2}{c}{} & \multicolumn{2}{c}{} & \multicolumn{2}{c}{} & \multicolumn{2}{c}{} & \multicolumn{2}{c}{} & \multicolumn{2}{c}{} & \multicolumn{2}{c}{} & \multicolumn{2}{c}{} & \multicolumn{2}{c|}{} & \multicolumn{2}{c}{} & \multicolumn{2}{c}{} \\
			\hline
			\multicolumn{2}{c|}{\multirow{2}[0]{*}{ROITransfomer \cite{roitrans}}} & \multicolumn{2}{c|}{\multirow{2}[0]{*}{Infrared}} & \multicolumn{2}{c}{\multirow{2}[0]{*}{76.1 }} & \multicolumn{2}{c}{\multirow{2}[0]{*}{51.7 }} & \multicolumn{2}{c}{\multirow{2}[0]{*}{51.9}} & \multicolumn{2}{c}{\multirow{2}[0]{*}{71.2}} & \multicolumn{2}{c}{\multirow{2}[0]{*}{64.3 }} & \multicolumn{2}{c}{\multirow{2}[0]{*}{70.7}} & \multicolumn{2}{c}{\multirow{2}[0]{*}{46.9 }} & \multicolumn{2}{c}{\multirow{2}[0]{*}{83.3}} & \multicolumn{2}{c|}{\multirow{2}[0]{*}{28.3 }} & \multicolumn{2}{c}{\multirow{2}[0]{*}{24.6 }} & \multicolumn{2}{c}{\multirow{2}[0]{*}{60.5 }} \\
			\multicolumn{2}{c|}{} & \multicolumn{2}{c|}{} & \multicolumn{2}{c}{} & \multicolumn{2}{c}{} & \multicolumn{2}{c}{} & \multicolumn{2}{c}{} & \multicolumn{2}{c}{} & \multicolumn{2}{c}{} & \multicolumn{2}{c}{} & \multicolumn{2}{c}{} & \multicolumn{2}{c|}{} & \multicolumn{2}{c}{} & \multicolumn{2}{c}{} \\
			\multicolumn{2}{c|}{\multirow{2}[0]{*}{Faster-RCNN \cite{fast}}} & \multicolumn{2}{c|}{\multirow{2}[0]{*}{Infrared}} & \multicolumn{2}{c}{\multirow{2}[0]{*}{71.6 }} & \multicolumn{2}{c}{\multirow{2}[0]{*}{49.1 }} & \multicolumn{2}{c}{\multirow{2}[0]{*}{49.2}} & \multicolumn{2}{c}{\multirow{2}[0]{*}{68.1}} & \multicolumn{2}{c}{\multirow{2}[0]{*}{57.0 }} & \multicolumn{2}{c}{\multirow{2}[0]{*}{66.5}} & \multicolumn{2}{c}{\multirow{2}[0]{*}{35.6 }} & \multicolumn{2}{c}{\multirow{2}[0]{*}{71.6}} & \multicolumn{2}{c|}{\multirow{2}[0]{*}{29.5 }} & \multicolumn{2}{c}{\multirow{2}[0]{*}{55.4 }} & \multicolumn{2}{c}{\multirow{2}[0]{*}{21.6 }} \\
			\multicolumn{2}{c|}{} & \multicolumn{2}{c|}{} & \multicolumn{2}{c}{} & \multicolumn{2}{c}{} & \multicolumn{2}{c}{} & \multicolumn{2}{c}{} & \multicolumn{2}{c}{} & \multicolumn{2}{c}{} & \multicolumn{2}{c}{} & \multicolumn{2}{c}{} & \multicolumn{2}{c|}{} & \multicolumn{2}{c}{} & \multicolumn{2}{c}{} \\
			\multicolumn{2}{c|}{\multirow{2}[0]{*}{Oriented R-CNN \cite{orcnn}}} & \multicolumn{2}{c|}{\multirow{2}[0]{*}{Infrared}} & \multicolumn{2}{c}{\multirow{2}[0]{*}{77.0 }} & \multicolumn{2}{c}{\multirow{2}[0]{*}{55.0 }} & \multicolumn{2}{c}{\multirow{2}[0]{*}{47.5}} & \multicolumn{2}{c}{\multirow{2}[0]{*}{73.6}} & \multicolumn{2}{c}{\multirow{2}[0]{*}{63.2 }} & \multicolumn{2}{c}{\multirow{2}[0]{*}{72.2}} & \multicolumn{2}{c}{\multirow{2}[0]{*}{49.4 }} & \multicolumn{2}{c}{\multirow{2}[0]{*}{79.6}} & \multicolumn{2}{c|}{\multirow{2}[0]{*}{30.5 }} & \multicolumn{2}{c}{\multirow{2}[0]{*}{27.1 }} & \multicolumn{2}{c}{\multirow{2}[0]{*}{60.9 }} \\
			\multicolumn{2}{c|}{} & \multicolumn{2}{c|}{} & \multicolumn{2}{c}{} & \multicolumn{2}{c}{} & \multicolumn{2}{c}{} & \multicolumn{2}{c}{} & \multicolumn{2}{c}{} & \multicolumn{2}{c}{} & \multicolumn{2}{c}{} & \multicolumn{2}{c}{} & \multicolumn{2}{c|}{} & \multicolumn{2}{c}{} & \multicolumn{2}{c}{} \\
			\multicolumn{2}{c|}{\multirow{2}[0]{*}{$\text{S}^2\text{ANet}$ \cite{S2ANet}}} & \multicolumn{2}{c|}{\multirow{2}[0]{*}{Infrared}} & \multicolumn{2}{c}{\multirow{2}[0]{*}{73.0 }} & \multicolumn{2}{c}{\multirow{2}[0]{*}{39.2 }} & \multicolumn{2}{c}{\multirow{2}[0]{*}{41.9}} & \multicolumn{2}{c}{\multirow{2}[0]{*}{59.2}} & \multicolumn{2}{c}{\multirow{2}[0]{*}{32.3 }} & \multicolumn{2}{c}{\multirow{2}[0]{*}{65.6}} & \multicolumn{2}{c}{\multirow{2}[0]{*}{13.9 }} & \multicolumn{2}{c}{\multirow{2}[0]{*}{12}} & \multicolumn{2}{c|}{\multirow{2}[0]{*}{23.1 }} & \multicolumn{2}{c}{\multirow{2}[0]{*}{40.0 }} & \multicolumn{2}{c}{\multirow{2}[0]{*}{17.2 }} \\
			\multicolumn{2}{c|}{} & \multicolumn{2}{c|}{} & \multicolumn{2}{c}{} & \multicolumn{2}{c}{} & \multicolumn{2}{c}{} & \multicolumn{2}{c}{} & \multicolumn{2}{c}{} & \multicolumn{2}{c}{} & \multicolumn{2}{c}{} & \multicolumn{2}{c}{} & \multicolumn{2}{c|}{} & \multicolumn{2}{c}{} & \multicolumn{2}{c}{} \\
			\hline
			\multicolumn{2}{c|}{\multirow{2}[0]{*}{$\text{C}^2\text{Former}$ \cite{c2former}}} & \multicolumn{2}{c|}{\multirow{2}[0]{*}{RGB+Infrared}} & \multicolumn{2}{c}{\multirow{2}[0]{*}{76.7 }} & \multicolumn{2}{c}{\multirow{2}[0]{*}{52.0 }} & \multicolumn{2}{c}{\multirow{2}[0]{*}{59.8}} & \multicolumn{2}{c}{\multirow{2}[0]{*}{63.2}} & \multicolumn{2}{c}{\multirow{2}[0]{*}{48.0 }} & \multicolumn{2}{c}{\multirow{2}[0]{*}{68.7}} & \multicolumn{2}{c}{\multirow{2}[0]{*}{43.3 }} & \multicolumn{2}{c}{\multirow{2}[0]{*}{47}} & \multicolumn{2}{c|}{\multirow{2}[0]{*}{41.9 }} & \multicolumn{2}{c}{\multirow{2}[0]{*}{55.6 }} & \multicolumn{2}{c}{\multirow{2}[0]{*}{22.9 }} \\
			\multicolumn{2}{c|}{} & \multicolumn{2}{c|}{} & \multicolumn{2}{c}{} & \multicolumn{2}{c}{} & \multicolumn{2}{c}{} & \multicolumn{2}{c}{} & \multicolumn{2}{c}{} & \multicolumn{2}{c}{} & \multicolumn{2}{c}{} & \multicolumn{2}{c}{} & \multicolumn{2}{c|}{} & \multicolumn{2}{c}{} & \multicolumn{2}{c}{} \\
			\multicolumn{2}{c|}{\multirow{2}[0]{*}{CMAFF \cite{cmff}}} & \multicolumn{2}{c|}{\multirow{2}[0]{*}{RGB+Infrared}} & \multicolumn{2}{c}{\multirow{2}[0]{*}{81.7 }} & \multicolumn{2}{c}{\multirow{2}[0]{*}{58.8 }} & \multicolumn{2}{c}{\multirow{2}[0]{*}{68.7}} & \multicolumn{2}{c}{\multirow{2}[0]{*}{78.4}} & \multicolumn{2}{c}{\multirow{2}[0]{*}{68.5 }} & \multicolumn{2}{c}{\multirow{2}[0]{*}{76.3}} & \multicolumn{2}{c}{\multirow{2}[0]{*}{66.0 }} & \multicolumn{2}{c}{\multirow{2}[0]{*}{72.7}} & \multicolumn{2}{c|}{\multirow{2}[0]{*}{51.5 }} & \multicolumn{2}{c}{\multirow{2}[0]{*}{69.2 }} & \multicolumn{2}{c}{\multirow{2}[0]{*}{30.5 }} \\
			\multicolumn{2}{c|}{} & \multicolumn{2}{c|}{} & \multicolumn{2}{c}{} & \multicolumn{2}{c}{} & \multicolumn{2}{c}{} & \multicolumn{2}{c}{} & \multicolumn{2}{c}{} & \multicolumn{2}{c}{} & \multicolumn{2}{c}{} & \multicolumn{2}{c}{} & \multicolumn{2}{c|}{} & \multicolumn{2}{c}{} & \multicolumn{2}{c}{} \\
			\multicolumn{2}{c|}{\multirow{2}[0]{*}{DMM \cite{dmm}}} & \multicolumn{2}{c|}{\multirow{2}[0]{*}{RGB+Infrared}} & \multicolumn{2}{c}{\multirow{2}[0]{*}{84.2 }} & \multicolumn{2}{c}{\multirow{2}[0]{*}{65.7 }} & \multicolumn{2}{c}{\multirow{2}[0]{*}{72.3}} & \multicolumn{2}{c}{\multirow{2}[0]{*}{79.0}} & \multicolumn{2}{c}{\multirow{2}[0]{*}{72.5 }} & \multicolumn{2}{c}{\multirow{2}[0]{*}{78.8}} & \multicolumn{2}{c}{\multirow{2}[0]{*}{72.3 }} & \multicolumn{2}{c}{\multirow{2}[0]{*}{93.6}} & \multicolumn{2}{c|}{\multirow{2}[0]{*}{56.2 }} & \multicolumn{2}{c}{\multirow{2}[0]{*}{75.0 }} & \multicolumn{2}{c}{\multirow{2}[0]{*}{39.7 }} \\
			\multicolumn{2}{c|}{} & \multicolumn{2}{c|}{} & \multicolumn{2}{c}{} & \multicolumn{2}{c}{} & \multicolumn{2}{c}{} & \multicolumn{2}{c}{} & \multicolumn{2}{c}{} & \multicolumn{2}{c}{} & \multicolumn{2}{c}{} & \multicolumn{2}{c}{} & \multicolumn{2}{c|}{} & \multicolumn{2}{c}{} & \multicolumn{2}{c}{} \\
			\multicolumn{2}{c|}{\multirow{2}[0]{*}{Ours}} & \multicolumn{2}{c|}{\multirow{2}[0]{*}{RGB+Infrared}} & \multicolumn{2}{c}{\multirow{2}[0]{*}{85.3 }} & \multicolumn{2}{c}{\multirow{2}[0]{*}{\textbf{66.7 }}} & \multicolumn{2}{c}{\multirow{2}[0]{*}{\textbf{74.1}}} & \multicolumn{2}{c}{\multirow{2}[0]{*}{\textbf{80.9}}} & \multicolumn{2}{c}{\multirow{2}[0]{*}{73.0 }} & \multicolumn{2}{c}{\multirow{2}[0]{*}{\textbf{79.4}}} & \multicolumn{2}{c}{\multirow{2}[0]{*}{\textbf{74.3 }}} & \multicolumn{2}{c}{\multirow{2}[0]{*}{\textbf{94.1}}} & \multicolumn{2}{c|}{\multirow{2}[0]{*}{\textbf{56.4 }}} & \multicolumn{2}{c}{\multirow{2}[0]{*}{\textbf{76.0 }}} & \multicolumn{2}{c}{\multirow{2}[0]{*}{\textbf{40.6 }}} \\
			\multicolumn{2}{c|}{} & \multicolumn{2}{c|}{} & \multicolumn{2}{c}{} & \multicolumn{2}{c}{} & \multicolumn{2}{c}{} & \multicolumn{2}{c}{} & \multicolumn{2}{c}{} & \multicolumn{2}{c}{} & \multicolumn{2}{c}{} & \multicolumn{2}{c}{} & \multicolumn{2}{c|}{} & \multicolumn{2}{c}{} & \multicolumn{2}{c}{} \\
			\hline
		\end{tabular}%
	}
\end{table*}%

\subsection{Comparing with SOTA Methods}
\subsubsection{Comparison Results on the DroneVehicle dataset}
Table \Rmnum{1} shows the comparison results on the DroneVehicle dataset. Our DEPFusion  is compared with single-modality methods and multi-modality methods respectively. It could be observed that DEPFusion  has significant performance in terms of mAP@0.5 and mAP@0.5:0.95. 

Concretely, for single RGB modality methods, RoITransformer \cite{roitrans}  achieves best performance with mAP@0.5 of 61.6\% and mAP@0.5:0.95 of 32.9\%, which is surpassed by our DEPFusion  by 17.3 points and 18.3 points respectively. Moreover, DEPFusion  achieves obvious performance for the category detection, such as the result on Car with mAP@0.5 of 90.5\% and Bus with mAP@0.5 of 90.0\%. For single infrared modality methods, all of the detectors perform better than adopting RGB modality only due to the harmful effects of low-light RGB images.
DEPFusion achieves best performance with mAP@0.5 of 78.9\%, while surpassed by DTNet \cite{dtnet} with mAP@0.5:0.95 of 52.9\%. However, DEPFusion  outperforms DTNet most in category detection, such as 3.7 points gains on Truck. DEPFusion  also surpasses some of multispectral detectors. Comparing with $\text{C}^2\text{Former}$ \cite{c2former}, it obtains 4.7 points improvement with mAP@0.5 and 3.7 points improvement with mAP@0.5:0.95.
Comparing with TarDAL, it obtains 7.3 points improvement with mAP@0.5 and 8.1 points improvement with mAP@0.5:0.95.

Fig \ref{showdv} shows the visualization results on the DroneVehicle dataset. DEPFusion  is compared with $\text{C}^2\text{Former}$ and DMM in the low-light scenarios, as the regions marked by blue circles show. In the first scene, $\text{C}^2\text{Former}$ detects a bus-stop as car and bus. DMM doesn't have such false detection but it classifies the freight-car as car. By contrast, our DEPFusion  avoids these mistakes. In the second scene, it could be observed that DEPFusion  makes correct classification of a car at night. In the third scenarios, $\text{C}^2\text{Former}$ and DMM make false detection that regards oval flower beds as cars at night, while DEPFusion  avoids this mistake. The fourth and fifth scene show DEPFusion's accurate classification under extreme low-light condition. These results shows that our method achieve robustness detection and classification in the low-light scenarios.

\subsubsection{Comparison Results on the VEDAI dataset}
Table \Rmnum{2} shows the comparison results on the VEDAI dataset. It could be observed that DEPFusion  outperforms many of detectors with mAP@0.5 of 71.1\%. However, for the Car and Van, DEPFusion  is surpassed by YOLO-S \cite{yolos} with 95.5\% and DAGN \cite{dgan} with 78.5\%. These two detectors only adopts RGB images to achieve object detection, they are be impacted by the redundant information from multi-modality fusion mechanism. Nevertheless, DEPFusion  also surpasses them on Truck, Boat and Other categories. Comparing with challenging DMM \cite{dmm}, DEPFusion  obtains 0.9 points improvement with mAP@0.5 and obtains most 2 points improvement with mAP@0.5 on Boat in terms of specific categories.

DEPFusion  is evaluated on VEDAI dataset \cite{vedai} with SOTA methods, the visualization results are shown in Fig \ref{vedai}. It could be observed that DEPFusion  could achieve correct classification and detection. In the first scene,  DEPFusion  reduces mis-classification of the cars on the road, comparing with $\text{C}^2\text{Former}$. The second scene shows the false detection and mis-classification of SOTA methods. Our DEPFusion  avoids these mistakes successfully. The similar performances could be observed in the last scene.

\subsubsection{Comparison of Computational Complexity}
Table \Rmnum{3} shows the comparison results of computational complexity between DEPFusion  and SOTA methods. Notably, this experiment is conducted on a NVIDIA RTX 3090 GPU, and the size of test input tensor is 512 $\times$ 640 with batch size of 1. It could be observed that our method has 80.3 MB parameter
count and 108.3 GFLOPs (Giga Floating-Point Operations Per
second). Especially, comparing $\text{C}^2\text{Former}$,  it reduces about 32.2\% parameter
count and 6.3\% GFLOPs. Comparing with DMM, it only reduces 7.8 MB parameter
count and 0.5 GFLOPs. For the accuracy, DEPFusion outperforms SOTA methods with mAP@0.5 of 78.9\%, gaining 5.8 points, 4.7 points and 0.3 points respectively when it is compared with TSFADet, $\text{C}^2\text{Former}$ and DMM.

\begin{figure}[t]
	\centering
	\includegraphics[width=3.4in]{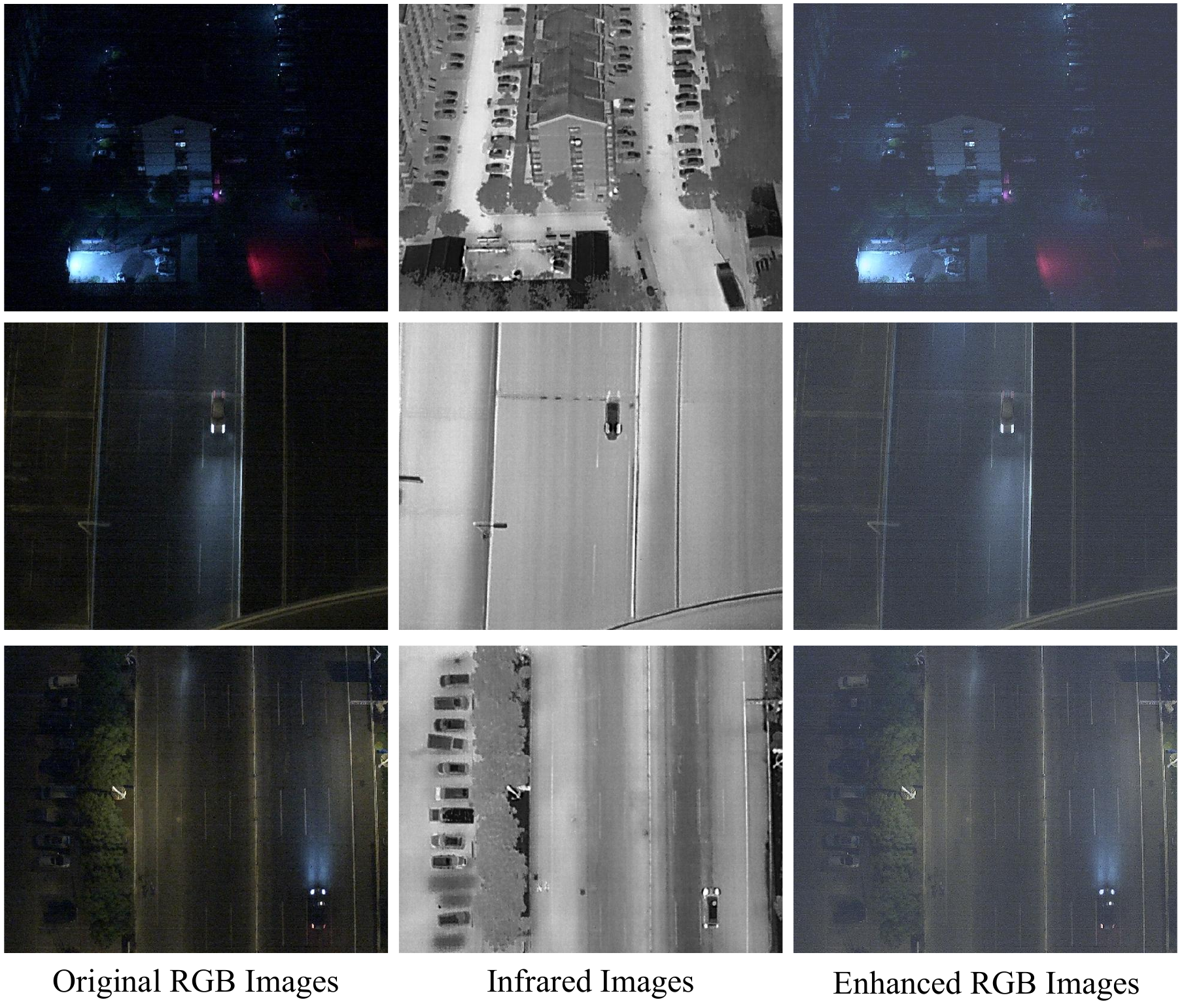}
	\caption{The comparison between original RGB images and enhanced RGB images, which are generated by the DDE module. \label{Enhance}
	}
\end{figure}
\begin{table}[t]
	\centering
	\small
	\caption{Ablation study on the designed module on the DroneVehicle dataset.\label{alsm}}
	\renewcommand{\arraystretch}{0.6}
	\setlength{\tabcolsep}{1.5mm}{
		\begin{tabular}{cc|cccc|cccc|cc|cc}
			\hline
			\multicolumn{2}{c|}{\multirow{4}[2]{*}{Models}} & \multicolumn{4}{c|}{\multirow{2}[1]{*}{DDE}} & \multicolumn{4}{c|}{\multirow{2}[1]{*}{Fusion}} & \multicolumn{4}{c}{\multirow{2}[1]{*}{mAP}} \\
			\multicolumn{2}{c|}{} & \multicolumn{4}{c|}{}         & \multicolumn{4}{c|}{}         & \multicolumn{4}{c}{} \\
			\cline{3-14}    \multicolumn{2}{c|}{} & \multicolumn{2}{c|}{\multirow{2}[1]{*}{CSWM}} & \multicolumn{2}{c|}{\multirow{2}[1]{*}{FDR}} & \multicolumn{2}{c|}{\multirow{2}[1]{*}{Z-order}} & \multicolumn{2}{c|}{\multirow{2}[1]{*}{PGMF}} & \multicolumn{2}{c|}{\multirow{2}[1]{*}{0.5}} & \multicolumn{2}{c}{\multirow{2}[1]{*}{0.5:0.95}} \\
			\multicolumn{2}{c|}{} & \multicolumn{2}{c|}{} & \multicolumn{2}{c|}{} & \multicolumn{2}{c|}{} & \multicolumn{2}{c|}{} & \multicolumn{2}{c|}{} & \multicolumn{2}{c}{} \\
			\hline
			\multicolumn{2}{c|}{\multirow{2}[1]{*}{model 1}} & \multicolumn{2}{c}{\multirow{2}[1]{*}{}} & \multicolumn{2}{c|}{\multirow{2}[1]{*}{}} & \multicolumn{2}{c}{\multirow{2}[1]{*}{\checkmark}} & \multicolumn{2}{c|}{\multirow{2}[1]{*}{}} & \multicolumn{2}{c|}{\multirow{2}[1]{*}{77.8 }} & \multicolumn{2}{c}{\multirow{2}[1]{*}{49.7 }} \\
			\multicolumn{2}{c|}{} & \multicolumn{2}{c}{} & \multicolumn{2}{c|}{} & \multicolumn{2}{c}{} & \multicolumn{2}{c|}{} & \multicolumn{2}{c|}{} & \multicolumn{2}{c}{} \\
			\multicolumn{2}{c|}{\multirow{2}[0]{*}{model 2}} & \multicolumn{2}{c}{\multirow{2}[0]{*}{}} & \multicolumn{2}{c|}{\multirow{2}[0]{*}{}} & \multicolumn{2}{c}{\multirow{2}[0]{*}{}} & \multicolumn{2}{c|}{\multirow{2}[0]{*}{\checkmark}} & \multicolumn{2}{c|}{\multirow{2}[0]{*}{78.1 }} & \multicolumn{2}{c}{\multirow{2}[0]{*}{50.8 }} \\
			\multicolumn{2}{c|}{} & \multicolumn{2}{c}{} & \multicolumn{2}{c|}{} & \multicolumn{2}{c}{} & \multicolumn{2}{c|}{} & \multicolumn{2}{c|}{} & \multicolumn{2}{c}{} \\
			\multicolumn{2}{c|}{\multirow{2}[0]{*}{model 3}} & \multicolumn{2}{c}{\multirow{2}[0]{*}{\checkmark}} & \multicolumn{2}{c|}{\multirow{2}[0]{*}{}} & \multicolumn{2}{c}{\multirow{2}[0]{*}{\checkmark}} & \multicolumn{2}{c|}{\multirow{2}[0]{*}{}} & \multicolumn{2}{c|}{\multirow{2}[0]{*}{77.8 }} & \multicolumn{2}{c}{\multirow{2}[0]{*}{50.0 }} \\
			\multicolumn{2}{c|}{} & \multicolumn{2}{c}{} & \multicolumn{2}{c|}{} & \multicolumn{2}{c}{} & \multicolumn{2}{c|}{} & \multicolumn{2}{c|}{} & \multicolumn{2}{c}{} \\
			\multicolumn{2}{c|}{\multirow{2}[0]{*}{model 4}} & \multicolumn{2}{c}{\multirow{2}[0]{*}{}} & \multicolumn{2}{c|}{\multirow{2}[0]{*}{\checkmark}} & \multicolumn{2}{c}{\multirow{2}[0]{*}{\checkmark}} & \multicolumn{2}{c|}{\multirow{2}[0]{*}{}} & \multicolumn{2}{c|}{\multirow{2}[0]{*}{75.4 }} & \multicolumn{2}{c}{\multirow{2}[0]{*}{49.1 }} \\
			\multicolumn{2}{c|}{} & \multicolumn{2}{c}{} & \multicolumn{2}{c|}{} & \multicolumn{2}{c}{} & \multicolumn{2}{c|}{} & \multicolumn{2}{c|}{} & \multicolumn{2}{c}{} \\
			\multicolumn{2}{c|}{\multirow{2}[0]{*}{model 5}} & \multicolumn{2}{c}{\multirow{2}[0]{*}{\checkmark}} & \multicolumn{2}{c|}{\multirow{2}[0]{*}{\checkmark}} & \multicolumn{2}{c}{\multirow{2}[0]{*}{\checkmark}} & \multicolumn{2}{c|}{\multirow{2}[0]{*}{}} & \multicolumn{2}{c|}{\multirow{2}[0]{*}{78.3 }} & \multicolumn{2}{c}{\multirow{2}[0]{*}{50.7 }} \\
			\multicolumn{2}{c|}{} & \multicolumn{2}{c}{} & \multicolumn{2}{c|}{} & \multicolumn{2}{c}{} & \multicolumn{2}{c|}{} & \multicolumn{2}{c|}{} & \multicolumn{2}{c}{} \\
			\multicolumn{2}{c|}{\multirow{2}[0]{*}{model 6}} & \multicolumn{2}{c}{\multirow{2}[0]{*}{\checkmark}} & \multicolumn{2}{c|}{\multirow{2}[0]{*}{\checkmark}} & \multicolumn{2}{c}{\multirow{2}[0]{*}{}} & \multicolumn{2}{c|}{\multirow{2}[0]{*}{\checkmark}} & \multicolumn{2}{c|}{\multirow{2}[0]{*}{\textbf{78.9 }}} & \multicolumn{2}{c}{\multirow{2}[0]{*}{\textbf{51.2 }}} \\
			\multicolumn{2}{c|}{} & \multicolumn{2}{c}{} & \multicolumn{2}{c|}{} & \multicolumn{2}{c}{} & \multicolumn{2}{c|}{} & \multicolumn{2}{c|}{} & \multicolumn{2}{c}{} \\
			\hline
	\end{tabular}}
\end{table}
\subsubsection{The Enhancement Effect of DDE Module}
Fig \ref{Enhance} shows the enhancement effect of DDE module. Comparing with original RGB images, our enhanced RGB images obtained from DDE module could show the details in the dark regions. For example, on the right side of the first scene, the clustered cars are observed due to their clear details. 
Similarly, the clustered cars are observed clearly on the left side of the third scene.
Therefore, DDE module enhances the global brightness and recovers the texture-details of low-light RGB images, which improves the complementarity in the multi-modality fusion stage.

\begin{figure}[t]
	\centering
	\includegraphics[width=3.5in]{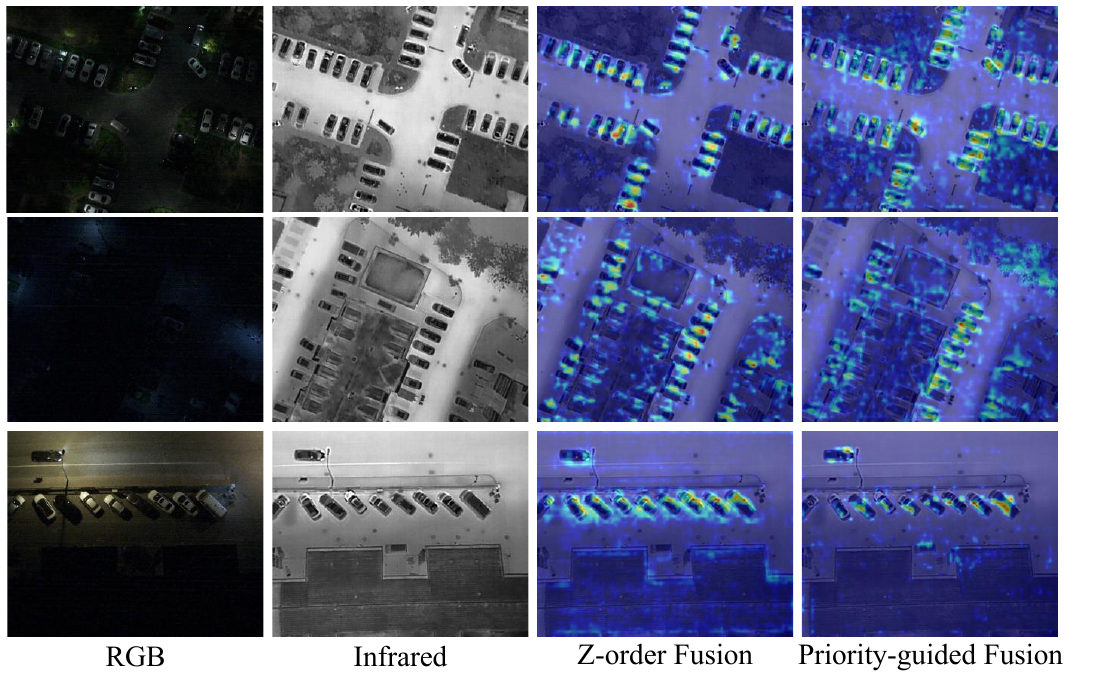}
	\caption{Comparison result about the Z-order Fusion and Priority-guided Fusion on the feature maps. 	\label{showhm}
	}
\end{figure}

\begin{table}[t]
	\centering
	\small
	\caption{Ablation study of the hyper-parameter in the CSWM block. LF and HF represents low-frequency component and high-frequency component respectively. \label{alscswm}}
	\renewcommand{\arraystretch}{0.6}
	\setlength{\tabcolsep}{1.4mm}{
		\begin{tabular}{cc|cccc|cccccc|cc|cc}
			\hline
			\multicolumn{2}{c|}{\multirow{4}[2]{*}{Models}} & \multicolumn{4}{c|}{\multirow{2}[1]{*}{Wavelet Basis}} & \multicolumn{6}{c|}{\multirow{2}[1]{*}{Enhanced Objects}} & \multicolumn{4}{c}{\multirow{2}[1]{*}{mAP}} \\
			\multicolumn{2}{c|}{} & \multicolumn{4}{c|}{}         & \multicolumn{6}{c|}{}                         & \multicolumn{4}{c}{} \\
			\cline{3-16}    \multicolumn{2}{c|}{} & \multicolumn{2}{c|}{\multirow{2}[1]{*}{Haar}} & \multicolumn{2}{c|}{\multirow{2}[1]{*}{Sym2}} & \multicolumn{2}{c|}{\multirow{2}[1]{*}{LF}} & \multicolumn{2}{c|}{\multirow{2}[1]{*}{HF}} & \multicolumn{2}{c|}{\multirow{2}[1]{*}{Together}} & \multicolumn{2}{c|}{\multirow{2}[1]{*}{0.5}} & \multicolumn{2}{c}{\multirow{2}[1]{*}{0.5:0.95}} \\
			\multicolumn{2}{c|}{} & \multicolumn{2}{c|}{} & \multicolumn{2}{c|}{} & \multicolumn{2}{c|}{} & \multicolumn{2}{c|}{} & \multicolumn{2}{c|}{} & \multicolumn{2}{c|}{} & \multicolumn{2}{c}{} \\
			\hline
			\multicolumn{2}{c|}{\multirow{2}[1]{*}{model 1}} & \multicolumn{2}{c}{\multirow{2}[1]{*}{}} & \multicolumn{2}{c|}{\multirow{2}[1]{*}{\checkmark}} & \multicolumn{2}{c}{\multirow{2}[1]{*}{}} & \multicolumn{2}{c}{\multirow{2}[1]{*}{}} & \multicolumn{2}{c|}{\multirow{2}[1]{*}{\checkmark}} & \multicolumn{2}{c|}{\multirow{2}[1]{*}{77.8 }} & \multicolumn{2}{c}{\multirow{2}[1]{*}{47.2 }} \\
			\multicolumn{2}{c|}{} & \multicolumn{2}{c}{} & \multicolumn{2}{c|}{} & \multicolumn{2}{c}{} & \multicolumn{2}{c}{} & \multicolumn{2}{c|}{} & \multicolumn{2}{c|}{} & \multicolumn{2}{c}{} \\
			\multicolumn{2}{c|}{\multirow{2}[0]{*}{model 2}} & \multicolumn{2}{c}{\multirow{2}[0]{*}{}} & \multicolumn{2}{c|}{\multirow{2}[0]{*}{\checkmark}} & \multicolumn{2}{c}{\multirow{2}[0]{*}{}} & \multicolumn{2}{c}{\multirow{2}[0]{*}{\checkmark}} & \multicolumn{2}{c|}{\multirow{2}[0]{*}{}} & \multicolumn{2}{c|}{\multirow{2}[0]{*}{75.6 }} & \multicolumn{2}{c}{\multirow{2}[0]{*}{46.6 }} \\
			\multicolumn{2}{c|}{} & \multicolumn{2}{c}{} & \multicolumn{2}{c|}{} & \multicolumn{2}{c}{} & \multicolumn{2}{c}{} & \multicolumn{2}{c|}{} & \multicolumn{2}{c|}{} & \multicolumn{2}{c}{} \\
			\multicolumn{2}{c|}{\multirow{2}[0]{*}{model 3}} & \multicolumn{2}{c}{\multirow{2}[0]{*}{}} & \multicolumn{2}{c|}{\multirow{2}[0]{*}{\checkmark}} & \multicolumn{2}{c}{\multirow{2}[0]{*}{\checkmark}} & \multicolumn{2}{c}{\multirow{2}[0]{*}{}} & \multicolumn{2}{c|}{\multirow{2}[0]{*}{}} & \multicolumn{2}{c|}{\multirow{2}[0]{*}{78.6 }} & \multicolumn{2}{c}{\multirow{2}[0]{*}{50.0 }} \\
			\multicolumn{2}{c|}{} & \multicolumn{2}{c}{} & \multicolumn{2}{c|}{} & \multicolumn{2}{c}{} & \multicolumn{2}{c}{} & \multicolumn{2}{c|}{} & \multicolumn{2}{c|}{} & \multicolumn{2}{c}{} \\
			\multicolumn{2}{c|}{\multirow{2}[0]{*}{model 4}} & \multicolumn{2}{c}{\multirow{2}[0]{*}{\checkmark}} & \multicolumn{2}{c|}{\multirow{2}[0]{*}{}} & \multicolumn{2}{c}{\multirow{2}[0]{*}{}} & \multicolumn{2}{c}{\multirow{2}[0]{*}{}} & \multicolumn{2}{c|}{\multirow{2}[0]{*}{\checkmark}} & \multicolumn{2}{c|}{\multirow{2}[0]{*}{77.8 }} & \multicolumn{2}{c}{\multirow{2}[0]{*}{50.1 }} \\
			\multicolumn{2}{c|}{} & \multicolumn{2}{c}{} & \multicolumn{2}{c|}{} & \multicolumn{2}{c}{} & \multicolumn{2}{c}{} & \multicolumn{2}{c|}{} & \multicolumn{2}{c|}{} & \multicolumn{2}{c}{} \\
			\multicolumn{2}{c|}{\multirow{2}[0]{*}{model 5}} & \multicolumn{2}{c}{\multirow{2}[0]{*}{\checkmark}} & \multicolumn{2}{c|}{\multirow{2}[0]{*}{}} & \multicolumn{2}{c}{\multirow{2}[0]{*}{}} & \multicolumn{2}{c}{\multirow{2}[0]{*}{\checkmark}} & \multicolumn{2}{c|}{\multirow{2}[0]{*}{}} & \multicolumn{2}{c|}{\multirow{2}[0]{*}{76.8 }} & \multicolumn{2}{c}{\multirow{2}[0]{*}{49.7 }} \\
			\multicolumn{2}{c|}{} & \multicolumn{2}{c}{} & \multicolumn{2}{c|}{} & \multicolumn{2}{c}{} & \multicolumn{2}{c}{} & \multicolumn{2}{c|}{} & \multicolumn{2}{c|}{} & \multicolumn{2}{c}{} \\
			\multicolumn{2}{c|}{\multirow{2}[0]{*}{model 6}} & \multicolumn{2}{c}{\multirow{2}[0]{*}{\checkmark}} & \multicolumn{2}{c|}{\multirow{2}[0]{*}{}} & \multicolumn{2}{c}{\multirow{2}[0]{*}{\checkmark}} & \multicolumn{2}{c}{\multirow{2}[0]{*}{}} & \multicolumn{2}{c|}{\multirow{2}[0]{*}{}} & \multicolumn{2}{c|}{\multirow{2}[0]{*}{\textbf{78.9 }}} & \multicolumn{2}{c}{\multirow{2}[0]{*}{\textbf{51.2 }}} \\
			\multicolumn{2}{c|}{} & \multicolumn{2}{c}{} & \multicolumn{2}{c|}{} & \multicolumn{2}{c}{} & \multicolumn{2}{c}{} & \multicolumn{2}{c|}{} & \multicolumn{2}{c|}{} & \multicolumn{2}{c}{} \\
			\hline
	\end{tabular}}
\end{table}%

\begin{table}[t]
	\centering
	\small
	\caption{Ablation study of scanning way in the CSWM block. CS and CSS represents Concatenate Scanning and Cross-Scale Scanning respectively. \label{alscswm2}}
	\renewcommand{\arraystretch}{0.6}
	\setlength{\tabcolsep}{2mm}{
		\begin{tabular}{cc|cccc|cc|cc}
			\hline
			\multicolumn{2}{c|}{\multirow{2}[1]{*}{Models}} & \multicolumn{2}{c}{\multirow{2}[1]{*}{CS}} & \multicolumn{2}{c|}{\multirow{2}[1]{*}{CSS}} & \multicolumn{2}{c|}{\multirow{2}[1]{*}{mAP@0.5}} & \multicolumn{2}{c}{\multirow{2}[1]{*}{mAP@0.5:0.95}} \\
			\multicolumn{2}{c|}{} & \multicolumn{2}{c}{} & \multicolumn{2}{c|}{} & \multicolumn{2}{c|}{} & \multicolumn{2}{c}{} \\
			\hline
			\multicolumn{2}{c|}{\multirow{2}[1]{*}{model 1}} & \multicolumn{2}{c}{\multirow{2}[1]{*}{\checkmark}} & \multicolumn{2}{c|}{\multirow{2}[1]{*}{}} & \multicolumn{2}{c|}{\multirow{2}[1]{*}{76.2}} & \multicolumn{2}{c}{\multirow{2}[1]{*}{47.9}} \\
			\multicolumn{2}{c|}{} & \multicolumn{2}{c}{} & \multicolumn{2}{c|}{} & \multicolumn{2}{c|}{} & \multicolumn{2}{c}{} \\
			\multicolumn{2}{c|}{\multirow{2}[1]{*}{model 2}} & \multicolumn{2}{c}{\multirow{2}[1]{*}{}} & \multicolumn{2}{c|}{\multirow{2}[1]{*}{\checkmark}} & \multicolumn{2}{c|}{\multirow{2}[1]{*}{78.9}} & \multicolumn{2}{c}{\multirow{2}[1]{*}{51.2}} \\
			\multicolumn{2}{c|}{} & \multicolumn{2}{c}{} & \multicolumn{2}{c|}{} & \multicolumn{2}{c|}{} & \multicolumn{2}{c}{} \\
			\hline
	\end{tabular}}
\end{table}%

\begin{table}[htbp]
	\centering
	\small
	\caption{Ablation study result on wavelet transformation level of 2D-DWT and kernel size combination in the CSWM block. \label{alslevelk}}
	\renewcommand{\arraystretch}{0.7}
	\setlength{\tabcolsep}{2.5mm}{
	\begin{tabular}{cccc|cc|cc}
		\hline
		\multicolumn{4}{c|}{\multirow{2}[1]{*}{Levels of 2D-DWT}} & \multicolumn{2}{c|}{\multirow{2}[1]{*}{mAP@0.5}} & \multicolumn{2}{c}{\multirow{2}[1]{*}{mAP@0.5:0.95}} \\
		\multicolumn{4}{c|}{}         & \multicolumn{2}{c|}{} & \multicolumn{2}{c}{} \\
		\hline
		\multicolumn{4}{c|}{\multirow{2}[1]{*}{1}} & \multicolumn{2}{c|}{\multirow{2}[1]{*}{76.3}} & \multicolumn{2}{c}{\multirow{2}[1]{*}{48.5}} \\
		\multicolumn{4}{c|}{}         & \multicolumn{2}{c|}{} & \multicolumn{2}{c}{} \\
		\multicolumn{4}{c|}{\multirow{2}[1]{*}{2}} & \multicolumn{2}{c|}{\multirow{2}[1]{*}{\textbf{78.9}}} & \multicolumn{2}{c}{\multirow{2}[1]{*}{\textbf{51.2}}} \\
		\multicolumn{4}{c|}{}         & \multicolumn{2}{c|}{} & \multicolumn{2}{c}{} \\
		\multicolumn{4}{c|}{\multirow{2}[1]{*}{3}} & \multicolumn{2}{c|}{\multirow{2}[1]{*}{78.9}} & \multicolumn{2}{c}{\multirow{2}[1]{*}{50.9}} \\
		\multicolumn{4}{c|}{}         & \multicolumn{2}{c|}{} & \multicolumn{2}{c}{} \\
		\multicolumn{4}{c|}{\multirow{2}[1]{*}{4}} & \multicolumn{2}{c|}{\multirow{2}[1]{*}{78.7}} & \multicolumn{2}{c}{\multirow{2}[1]{*}{50.6}} \\
		\multicolumn{4}{c|}{}         & \multicolumn{2}{c|}{} & \multicolumn{2}{c}{} \\
		\hline
		\multicolumn{4}{c|}{\multirow{2}[1]{*}{Kernel Size Combination}} & \multicolumn{2}{c|}{\multirow{2}[1]{*}{mAP@0.5}} & \multicolumn{2}{c}{\multirow{2}[1]{*}{mAP@0.5:0.95}} \\
		\multicolumn{4}{c|}{}         & \multicolumn{2}{c|}{} & \multicolumn{2}{c}{} \\
		\hline
		\multicolumn{4}{c|}{\multirow{2}[1]{*}{[5,7,9]}} & \multicolumn{2}{c|}{\multirow{2}[1]{*}{78.7}} & \multicolumn{2}{c}{\multirow{2}[1]{*}{50.0}} \\
		\multicolumn{4}{c|}{}         & \multicolumn{2}{c|}{} & \multicolumn{2}{c}{} \\
		\multicolumn{4}{c|}{\multirow{2}[1]{*}{[3,5,9]}} & \multicolumn{2}{c|}{\multirow{2}[1]{*}{78.5}} & \multicolumn{2}{c}{\multirow{2}[1]{*}{51.0}} \\
		\multicolumn{4}{c|}{}         & \multicolumn{2}{c|}{} & \multicolumn{2}{c}{} \\
		\multicolumn{4}{c|}{\multirow{2}[1]{*}{[3,7,9]}} & \multicolumn{2}{c|}{\multirow{2}[1]{*}{78.7}} & \multicolumn{2}{c}{\multirow{2}[1]{*}{50.8}} \\
		\multicolumn{4}{c|}{}         & \multicolumn{2}{c|}{} & \multicolumn{2}{c}{} \\
		\multicolumn{4}{c|}{\multirow{2}[1]{*}{[3,5,7]}} & \multicolumn{2}{c|}{\multirow{2}[1]{*}{\textbf{78.9}}} & \multicolumn{2}{c}{\multirow{2}[1]{*}{\textbf{51.2}}} \\
		\multicolumn{4}{c|}{}         & \multicolumn{2}{c|}{} & \multicolumn{2}{c}{} \\
		\hline
	\end{tabular}}

\end{table}%

\subsection{Ablation Study}
\subsubsection{Ablation Study on the Designed Modules}
The ablation study about the designed modules
is conducted and the result is shown in Table \Rmnum{4}. Comparing model 1 and model 2, the mAP@0.5 and mAP@0.5:0.95 improve about 0.3 points and 1.1 points when Z-order fusion is replaced with PGMF module, which demonstrates that the priority scanning could achieve more effective fusion than Z-order scanning. Comparing model 3, model 4 and model 5, it could be seen that introducing CSWM and FDR block only could not improve the metrics. In contrast, he mAP@0.5 and mAP@0.5:0.95 improve about 0.5 points and 1 points when both of them are introduced. This result demonstrates that enhancing the brightness and recovering the texture-details are important for detection. After adding PGMF module, the metrics improve to the optimum, which demonstrates that priority scanning fusion with enhanced images feature could achieve more correct detection.

\subsubsection{Ablation Study on CSWM Module}
The ablation study of the settings in CSWM block is conducted, as shown in Table \Rmnum{5} and Table \Rmnum{6}. For the wavelet basis, adopting Haar wavelet improve the metrics more than adopting Sym2 wavelet, because the latter brings more noise of the local details. For the enhanced objects, only the LF takes part in the enhancement could achieve good effect, no matter what types of wavelet basis. This is because the HF contains less important texture-details about the images, introducing HF into enhancement brings more redundant information. In addition, only enhancing HF could not enhance the global brightness of images.
Therefore, the texture-details are enhanced by our designed FDR block additionally.

Table \Rmnum{6} shows the effect of designed Cross-Scale Scanning mechanism (CSS). Notably, the Concatenate Scanning (CS) means that combining simple Z-order sequences as input. From result, it could be seen that introducing CSS improves mAP@0.5 with 2.7 points and mAP@0.5:0.95 with 3.3 points. This is because CSS learns multi-scale features and focuses on local brightness distribution between difference scales. Therefore, the global brightness is enhanced after the scanning and combination of Mamba.

The effect brought by the transformation level of 2D-DWT in CSWM is evaluated in Table \Rmnum{7}. It could be observed that the values of two metric are approximate when level=2 and level=3, with mAP@0.5 of 78.9\% and mAP@0.5:0.95 of 51.0\%. The metric values decline as the level=4. Therefore, from the perspective of computational costs, the transformation level is set to 2 during training.

In addition, Table \Rmnum{7} shows the effect of kernel size combination in the CSWM block. The multi-scale feature of low-frequency component is obtained by three different sizes of convolution kernel. The results show that the span of size variation brings small effects on mAP@0.5 and mAP@0.5:0.95. These small effects occur due to the low multi-scale complementary caused by obvious size differences. 

\subsubsection{Visualization of fusion feature maps}
To evaluate our priority scanning, the visualization of fusion feature maps are shown in Fig \ref{showhm}. It could be observed that the degree of focus is located on background easily under Z-order fusion mechanism. By contrast, our priority scanning fusion shifts the focus of attention to the instances, suppressing the interference from background information.

\subsubsection{Ablation Study on Fusion Ways}
Fig \ref{fuseway} shows four different ways of fusion priority sequences. Additionally, in the Method Section, we have proved that the fusion sequence satisfies the format of Fig. 9 (d). Therefore, the ablation study about the fusion ways is conducted, whose results could be seen in Table \Rmnum{8}. The results of way (b) and way (c) prove that the locations of high priority tokens have an impact on accuracy, which shows that the early tokens are more important than late tokens because the former make better parameter renewing of $A$ matrix if related information is provided. For way (a), when scanning arrives at the end of sequences, the contributions of high priority tokens are decayed. The parameter renewing of $A$ matrix is affected by early tokens which contain background information. These two factors lead to unsteady accuracy, like mAP@0.5:0.95 of 48.9\%. Following way (d) from our proof, DEPFusion obtains mAP@0.5 with 78.9\% and mAP@0.5:0.95 of 51.2\%.

Notably, the theory of PGMF module is proved from the perspective of contribution's limit. Therefore, for an infinitely long sequence, the advantage of priority-guided serialization may be obvious. However, the sequences in real scenes are finite length almost. Therefore, our future work would explore the theory's scope of application further.

\begin{figure}[t]
	\centering
	\includegraphics[width=3.2in]{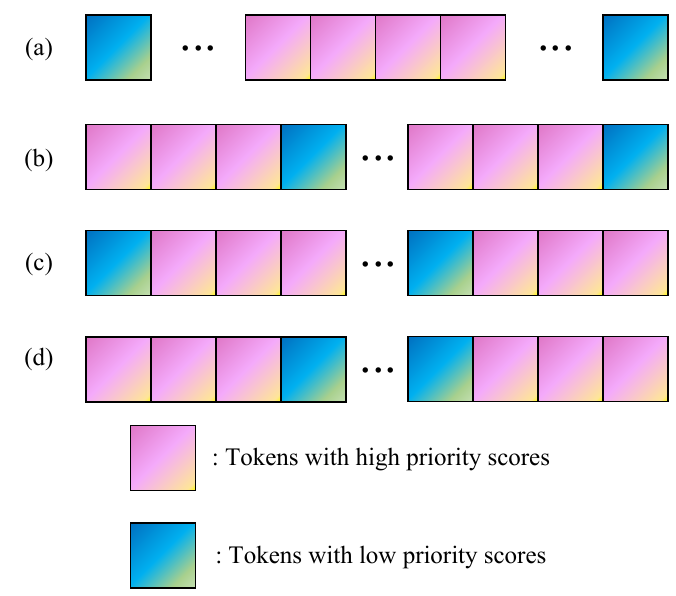}
	\caption{Different ways of fusion priority sequences. (a) means $
		\overline{x_{vp}}+x_{ip} $; (b) means $
		x_{vp}+x_{ip} $; (c) means $
		\overline{x_{vp}}+\overline{x_{ip}} $ and (d) means $x_{vp}+\overline{x_{ip}} $.\label{sort}
	}
\end{figure}

\begin{table}[t]
	\centering
	\small
	\caption{Ablation study result on different fusion ways. \label{fuseway}}
	\renewcommand{\arraystretch}{0.8}
	\setlength{\tabcolsep}{2.5mm}{
	\begin{tabular}{cc|cc|cc}
		\hline
		\multicolumn{2}{c|}{\multirow{2}[1]{*}{Fusing Way}} & \multicolumn{2}{c|}{\multirow{2}[1]{*}{mAP@0.5}} & \multicolumn{2}{c}{\multirow{2}[1]{*}{mAP@0.5:0.95}} \\
		\multicolumn{2}{c|}{} & \multicolumn{2}{c|}{} & \multicolumn{2}{c}{} \\
		\hline
		\multicolumn{2}{c|}{\multirow{2}[0]{*}{$\overline{x_{vp}}+x_{ip} $ (a)}} & \multicolumn{2}{c|}{\multirow{2}[0]{*}{76.6}} & \multicolumn{2}{c}{\multirow{2}[0]{*}{48.9}} \\
		\multicolumn{2}{c|}{} & \multicolumn{2}{c|}{} & \multicolumn{2}{c}{} \\
		\multicolumn{2}{c|}{\multirow{2}[0]{*}{$x_{vp}+x_{ip} $ (b)}} & \multicolumn{2}{c|}{\multirow{2}[0]{*}{76.4}} & \multicolumn{2}{c}{\multirow{2}[0]{*}{50.6}} \\
		\multicolumn{2}{c|}{} & \multicolumn{2}{c|}{} & \multicolumn{2}{c}{} \\
		\multicolumn{2}{c|}{\multirow{2}[0]{*}{$\overline{x_{vp}}+\overline{x_{ip}} $ (c)}} & \multicolumn{2}{c|}{\multirow{2}[0]{*}{75.8}} & \multicolumn{2}{c}{\multirow{2}[0]{*}{49.7}} \\
		\multicolumn{2}{c|}{} & \multicolumn{2}{c|}{} & \multicolumn{2}{c}{} \\
		\multicolumn{2}{c|}{\multirow{2}[0]{*}{$x_{vp}+\overline{x_{ip}} $ (d)}} & \multicolumn{2}{c|}{\multirow{2}[0]{*}{\textbf{78.9}}} & \multicolumn{2}{c}{\multirow{2}[0]{*}{\textbf{51.2}}} \\
		\multicolumn{2}{c|}{} & \multicolumn{2}{c|}{} & \multicolumn{2}{c}{} \\
		\hline
	\end{tabular}%
}
\end{table}%

\section{Conclusion}
In this article, DEPFusion  is proposed with DDE module and PGMF module. DDE contains CSWM and FDR block to enhance the global brightness and recover texture-details of low-light RGB images. PGMF introduces Priority Scanning to enhance the local target modeling and achieve feature alignment. To our best knowledge, PGMF is the first approach to improve Mamba mechanism for multi-modality fusion. The experiment results shows that DEPFusion  performs well on object detection and achieves lightweight computation.

\bibliographystyle{IEEEtran}
\bibliography{New_IEEEtran_how-to}

\begin{thebibliography}{10}
\providecommand{\url}[1]{#1}
\csname url@samestyle\endcsname
\providecommand{\newblock}{\relax}
\providecommand{\bibinfo}[2]{#2}
\providecommand{\BIBentrySTDinterwordspacing}{\spaceskip=0pt\relax}
\providecommand{\BIBentryALTinterwordstretchfactor}{4}
\providecommand{\BIBentryALTinterwordspacing}{\spaceskip=\fontdimen2\font plus
\BIBentryALTinterwordstretchfactor\fontdimen3\font minus
  \fontdimen4\font\relax}
\providecommand{\BIBforeignlanguage}[2]{{%
\expandafter\ifx\csname l@#1\endcsname\relax
\typeout{** WARNING: IEEEtran.bst: No hyphenation pattern has been}%
\typeout{** loaded for the language `#1'. Using the pattern for}%
\typeout{** the default language instead.}%
\else
\language=\csname l@#1\endcsname
\fi
#2}}
\providecommand{\BIBdecl}{\relax}
\BIBdecl

\bibitem{bridging}
T.~Zhou, J.~Chen, Y.~Shi, K.~Jiang, M.~Yang, and D.~Yang, ``Bridging the view
  disparity between radar and camera features for multi-modal fusion 3d object
  detection,'' \emph{IEEE Trans. Intell. Veh.}, vol.~8, no.~2, pp. 1523--1535,
  2023.

\bibitem{sun}
S.~Tao, Y.~Shengqi, L.~Haiying, G.~Jason, D.~Lixia, and L.~Lida, ``Mis-yolov8:
  An improved algorithm for detecting small objects in uav aerial photography
  based on yolov8,'' \emph{IEEE Trans. Instrum. Meas.}, vol.~74, pp. 1--12,
  2025.

\bibitem{yolov8}


\bibitem{xiao}
Y.~Xiao, T.~Xu, X.~Yu, Y.~Fang, and J.~Li, ``A lightweight fusion strategy with
  enhanced interlayer feature correlation for small object detection,''
  \emph{IEEE Trans. Geosci. Remote Sens.}, vol.~62, pp. 1--11, 2024.

\bibitem{Lin}
H.~Lin, N.~Li, P.~Yao, K.~Dong, Y.~Guo, D.~Hong, Y.~Zhang, and C.~Wen,
  ``Generalization-enhanced few-shot object detection in remote sensing,''
  \emph{IEEE Trans. Circuits Syst. Video Technol.}, vol.~35, no.~6, pp.
  5445--5460, 2025.

\bibitem{Liang}
X.~Liang, J.~Zhang, L.~Zhuo, Y.~Li, and Q.~Tian, ``Small object detection in
  unmanned aerial vehicle images using feature fusion and scaling-based single
  shot detector with spatial context analysis,'' \emph{IEEE Trans. Circuits
  Syst. Video Technol.}, vol.~30, no.~6, pp. 1758--1770, 2020.

\bibitem{Yu}
Y.~Yu, K.~Zhang, X.~Wang, N.~Wang, and X.~Gao, ``An adaptive region proposal
  network with progressive attention propagation for tiny person detection from
  uav images,'' \emph{IEEE Trans. Circuits Syst. Video Technol.}, vol.~34,
  no.~6, pp. 4392--4406, 2024.

\bibitem{tardal}
J.~Liu, X.~Fan, Z.~Huang, G.~Wu, R.~Liu, W.~Zhong, and Z.~Luo, ``Target-aware
  dual adversarial learning and a multi-scenario multi-modality benchmark to
  fuse infrared and visible for object detection,'' in \emph{Proc. IEEE/CVF
  Conf. Comput. Vis. Pattern Recognit.}, 2022, pp. 5802--5811.

\bibitem{superyolo}
J.~Zhang, J.~Lei, W.~Xie, Z.~Fang, Y.~Li, and Q.~Du, ``Superyolo: Super
  resolution assisted object detection in multimodal remote sensing imagery,''
  \emph{IEEE Trans. Geosci. Remote Sens.}, vol.~61, pp. 1--15, 2023.

\bibitem{detfusion}
Y.~Sun, B.~Cao, P.~Zhu, and Q.~Hu, ``Detfusion: A detection-driven infrared and
  visible image fusion network,'' in \emph{ACM Int. Conf. Multimedia}, 2022,
  pp. 4003--4011.

\bibitem{dronevehicle}
------, ``Drone-based rgb-infrared cross-modality vehicle detection via
  uncertainty-aware learning,'' \emph{IEEE Trans. Circuits Syst. Video
  Technol.}, vol.~32, no.~10, pp. 6700--6713, 2022.

\bibitem{tracking}
P.~Zhang, J.~Zhao, D.~Wang, H.~Lu, and X.~Ruan, ``Visible-thermal uav tracking:
  A large-scale benchmark and new baseline,'' in \emph{Proc. IEEE/CVF Conf.
  Comput. Vis. Pattern Recognit.}, 2022, pp. 8886--8895.

\bibitem{c2former}
M.~Yuan and X.~Wei, ``C$^2$former: Calibrated and complementary transformer for
  rgb-infrared object detection,'' \emph{IEEE Trans. Geosci. Remote Sens.},
  vol.~62, pp. 1--12, 2024.

\bibitem{dmm}
M.~Zhou, T.~Li, C.~Qiao, D.~Xie, G.~Wang, N.~Ruan, L.~Mei, Y.~Yang, and H.~T.
  Shen, ``Dmm: Disparity-guided multispectral mamba for oriented object
  detection in remote sensing,'' \emph{IEEE Trans. Geosci. Remote Sens.}, 2025.

\bibitem{multi}
B.~Cao, Y.~Sun, P.~Zhu, and Q.~Hu, ``Multi-modal gated mixture of
  local-to-global experts for dynamic image fusion,'' in \emph{Proc. IEEE/CVF
  Conf. Comput. Vis. Pattern Recognit.}, 2023, pp. 23\,555--23\,564.

\bibitem{tsr}
M.~Yuan, Y.~Wang, and X.~Wei, ``Translation, scale and rotation: Cross-modal
  alignment meets rgb-infrared vehicle detection,'' in \emph{Eur. Conf. Comput.
  Vis.}\hskip 1em plus 0.5em minus 0.4em\relax Springer, 2022, pp. 509--525.

\bibitem{removal}
T.~Zhao, M.~Yuan, F.~Jiang, N.~Wang, and X.~Wei, ``Removal then selection: A
  coarse-to-fine fusion perspective for rgb-infrared object detection,''
  \emph{arXiv e-prints}, pp. arXiv--2401, 2024.

\bibitem{fourll}
H.~Feng, L.~Wang, Y.~Wang, and H.~Huang, ``Learnability enhancement for
  low-light raw denoising: Where paired real data meets noise modeling,'' in
  \emph{Proceedings of the 30th ACM International Conference on Multimedia},
  2022, pp. 1436--1444.

\bibitem{diffll}
H.~Jiang, A.~Luo, H.~Fan, S.~Han, and S.~Liu, ``Low-light image enhancement
  with wavelet-based diffusion models,'' \emph{ACM Trans. Graphics}, vol.~42,
  no.~6, pp. 1--14, 2023.

\bibitem{lowlightaaai}
H.~Zhou, W.~Dong, X.~Liu, Y.~Zhang, G.~Zhai, and J.~Chen, ``Low-light image
  enhancement via generative perceptual priors,'' in \emph{Proc.AAAI Conf.
  Artif. Intell.}, vol.~39, no.~10, 2025, pp. 10\,752--10\,760.

\bibitem{wavemamba}
W.~Zou, H.~Gao, W.~Yang, and T.~Liu, ``Wave-mamba: Wavelet state space model
  for ultra-high-definition low-light image enhancement,'' in \emph{ACM Int.
  Conf. Multimedia}, 2024, pp. 1534--1543.

\bibitem{mamba}
A.~Gu and T.~Dao, ``Mamba: Linear-time sequence modeling with selective state
  spaces,'' \emph{arXiv preprint arXiv:2312.00752}, 2023.

\bibitem{vmamba}
Y.~Liu, Y.~Tian, Y.~Zhao, H.~Yu, L.~Xie, Y.~Wang, Q.~Ye, J.~Jiao, and Y.~Liu,
  ``Vmamba: Visual state space model,'' \emph{Adv. Neural Inf. Process. Syst.},
  vol.~37, pp. 103\,031--103\,063, 2024.

\bibitem{vim}
L.~Zhu, B.~Liao, Q.~Zhang, X.~Wang, W.~Liu, and X.~Wang, ``Vision mamba:
  Efficient visual representation learning with bidirectional state space
  model,'' \emph{arXiv preprint arXiv:2401.09417}, 2024.

\bibitem{spectralmamba}
J.~Yao, D.~Hong, C.~Li, and J.~Chanussot, ``Spectralmamba: Efficient mamba for
  hyperspectral image classification,'' \emph{arXiv preprint arXiv:2404.08489},
  2024.

\bibitem{s2mamba}
G.~Wang, X.~Zhang, Z.~Peng, T.~Zhang, and L.~Jiao, ``S 2 mamba: A
  spatial-spectral state space model for hyperspectral image classification,''
  \emph{IEEE Trans. Geosci. Remote Sens.}, 2025.

\bibitem{rs3mamba}
X.~Ma, X.~Zhang, and M.-O. Pun, ``Rs 3 mamba: Visual state space model for
  remote sensing image semantic segmentation,'' \emph{IEEE Geosci. Remote Sens.
  Lett.}, vol.~21, pp. 1--5, 2024.

\bibitem{vmunet}
J.~Ruan, J.~Li, and S.~Xiang, ``Vm-unet: Vision mamba unet for medical image
  segmentation,'' \emph{arXiv preprint arXiv:2402.02491}, 2024.

\bibitem{celoss}
C.~E. Shannon, ``A mathematical theory of communication,'' \emph{The Bell
  system technical journal}, vol.~27, no.~3, pp. 379--423, 1948.

\bibitem{l1loss}
S.~Razakarivony and F.~Jurie, ``Vehicle detection in aerial imagery: A small
  target detection benchmark,'' \emph{Journal of Visual Communication and Image
  Representation}, vol.~34, pp. 187--203, 2016.

\bibitem{retinanet}
T.-Y. Lin, P.~Goyal, R.~Girshick, K.~He, and P.~Doll{\'a}r, ``Focal loss for
  dense object detection,'' in \emph{Proc. IEEE/CVF Conf. Comput. Vis. Pattern
  Recognit.}, 2017, pp. 2980--2988.

\bibitem{r3det}
X.~Yang, J.~Yan, Z.~Feng, and T.~He, ``R3det: Refined single-stage detector
  with feature refinement for rotating object,'' in \emph{Proc.AAAI Conf.
  Artif. Intell.}, vol.~35, no.~4, 2021, pp. 3163--3171.

\bibitem{kfiou}
X.~Yang, Y.~Zhou, G.~Zhang, J.~Yang, W.~Wang, J.~Yan, X.~Zhang, and Q.~Tian,
  ``The kfiou loss for rotated object detection,'' \emph{arXiv preprint
  arXiv:2201.12558}, 2022.

\bibitem{roitrans}
J.~Ding, N.~Xue, Y.~Long, G.-S. Xia, and Q.~Lu, ``Learning roi transformer for
  oriented object detection in aerial images,'' in \emph{Proc. IEEE/CVF Conf.
  Comput. Vis. Pattern Recognit.}, 2019, pp. 2849--2858.

\bibitem{YOLOV5}
G.~Jocher, A.~Stoken, J.~Borovec, L.~Changyu, A.~Hogan, L.~Diaconu,
  J.~Poznanski, L.~Yu, P.~Rai, R.~Ferriday \emph{et~al.}, ``ultralytics/yolov5:
  v3. 0,'' \emph{Zenodo}, 2020.

\bibitem{orcnn}
X.~Xie, G.~Cheng, J.~Wang, X.~Yao, and J.~Han, ``Oriented r-cnn for object
  detection,'' in \emph{Proc. IEEE/CVF Conf. Comput. Vis. Pattern Recognit.},
  2021, pp. 3520--3529.

\bibitem{PKINet-S}
X.~Cai, Q.~Lai, Y.~Wang, W.~Wang, Z.~Sun, and Y.~Yao, ``Poly kernel inception
  network for remote sensing detection,'' in \emph{Proc. IEEE/CVF Conf. Comput.
  Vis. Pattern Recognit.}, 2024, pp. 27\,706--27\,716.

\bibitem{S2ANet}
J.~Han, J.~Ding, J.~Li, and G.-S. Xia, ``Align deep features for oriented
  object detection,'' \emph{IEEE Trans. Geosci. Remote Sens.}, vol.~60, pp.
  1--11, 2021.

\bibitem{redet}
J.~Han, J.~Ding, N.~Xue, and G.-S. Xia, ``Redet: A rotation-equivariant
  detector for aerial object detection,'' in \emph{Proc. IEEE/CVF Conf. Comput.
  Vis. Pattern Recognit.}, 2021, pp. 2786--2795.

\bibitem{GWD}
X.~Yang, J.~Yan, Q.~Ming, W.~Wang, X.~Zhang, and Q.~Tian, ``Rethinking rotated
  object detection with gaussian wasserstein distance loss,'' in \emph{Proc.
  Int. Conf. Mach. Learn.}\hskip 1em plus 0.5em minus 0.4em\relax PMLR, 2021,
  pp. 11\,830--11\,841.

\bibitem{dtnet}
N.~Zhang, Y.~Liu, H.~Liu, T.~Tian, J.~Ma, and J.~Tian, ``Dtnet: A specialized
  dual-tuning network for infrared vehicle detection in aerial images,''
  \emph{IEEE Trans. Geosci. Remote Sens.}, vol.~62, pp. 1--15, 2024.

\bibitem{GLFNet}
X.~Kang, H.~Yin, and P.~Duan, ``Global--local feature fusion network for
  visible--infrared vehicle detection,'' \emph{IEEE Geosci. Remote Sens.
  Lett.}, vol.~21, pp. 1--5, 2024.

\bibitem{AR-CNN}
L.~Zhang, Z.~Liu, X.~Zhu, Z.~Song, X.~Yang, Z.~Lei, and H.~Qiao, ``Weakly
  aligned feature fusion for multimodal object detection,'' \emph{IEEE Trans.
  Neural Networks Learn. Syst.}, 2021.

\bibitem{MBNet}
K.~Zhou, L.~Chen, and X.~Cao, ``Improving multispectral pedestrian detection by
  addressing modality imbalance problems,'' in \emph{Eur. Conf. Comput.
  Vis.}\hskip 1em plus 0.5em minus 0.4em\relax Springer, 2020, pp. 787--803.

\bibitem{CIAN}
L.~Zhang, Z.~Liu, S.~Zhang, X.~Yang, H.~Qiao, K.~Huang, and A.~Hussain,
  ``Cross-modality interactive attention network for multispectral pedestrian
  detection,'' \emph{Information Fusion}, vol.~50, pp. 20--29, 2019.

\bibitem{TFDet}
X.~Zhang, X.~Zhang, J.~Wang, J.~Ying, Z.~Sheng, H.~Yu, C.~Li, and H.-L. Shen,
  ``Tfdet: Target-aware fusion for rgb-t pedestrian detection,'' \emph{IEEE
  Trans. Neural Networks Learn. Syst.}, vol.~36, no.~7, pp. 13\,276--13\,290,
  2025.

\bibitem{AFFNet}
Z.~Chen, W.~Xiang, Z.~Lin, K.~Yang, Y.~Liu, and Z.~Shi, ``Alignment-assisted
  frequency fusion network for rgb-infrared vehicle detection,''
  \emph{Neurocomputing}, vol. 647, p. 130505, 2025.

\bibitem{vedai}
S.~Razakarivony and F.~Jurie, ``Vehicle detection in aerial imagery : A small
  target detection benchmark,'' \emph{Journal of Visual Communication and Image
  Representation}, vol.~34, pp. 187--203, 2016.

\bibitem{mmdetection}
K.~Chen, J.~Wang, J.~Pang, Y.~Cao, Y.~Xiong, X.~Li, S.~Sun, W.~Feng, Z.~Liu,
  J.~Xu, Z.~Zhang, D.~Cheng, C.~Zhu, T.~Cheng, Q.~Zhao, B.~Li, X.~Lu, R.~Zhu,
  Y.~Wu, J.~Dai, J.~Wang, J.~Shi, W.~Ouyang, C.~C. Loy, and D.~Lin,
  ``{MMDetection}: Open mmlab detection toolbox and benchmark,'' \emph{arXiv
  preprint arXiv:1906.07155}, 2019.

\bibitem{mmrotate}
Y.~Zhou, X.~Yang, G.~Zhang, J.~Wang, Y.~Liu, L.~Hou, X.~Jiang, X.~Liu, J.~Yan,
  C.~Lyu, W.~Zhang, and K.~Chen, ``Mmrotate: A rotated object detection
  benchmark using pytorch,'' in \emph{ACM Int. Conf. Multimedia}, 2022.

\bibitem{yolos}
A.~Betti and M.~Tucci, ``Yolo-s: a lightweight and accurate yolo-like network
  for small target detection in aerial imagery,'' \emph{Sensors}, vol.~23,
  no.~4, p. 1865, 2023.

\bibitem{yolov3}
J.~Redmon and A.~Farhadi, ``Yolov3: An incremental improvement,'' \emph{arXiv
  preprint arXiv:1804.02767}, 2018.

\bibitem{fast}
R.~Girshick, ``Fast r-cnn,'' in \emph{Proc. IEEE/CVF Conf. Comput. Vis. Pattern
  Recognit.}, 2015, pp. 1440--1448.

\bibitem{cmff}
F.~Qingyun and W.~Zhaokui, ``Cross-modality attentive feature fusion for object
  detection in multispectral remote sensing imagery,'' \emph{Pattern
  Recognit.}, vol. 130, p. 108786, 2022.

\bibitem{dgan}
Z.~Zhang, Y.~Liu, T.~Liu, Z.~Lin, and S.~Wang, ``Dagn: A real-time uav remote
  sensing image vehicle detection framework,'' \emph{IEEE Geosci. Remote Sens.
  Lett.}, vol.~17, no.~11, pp. 1884--1888, 2019.

\end{thebibliography}

\end{document}